\documentclass[journal]{IEEEtran}
\usepackage{cite}
\usepackage{placeins}
\usepackage{amsmath,amssymb,amsfonts}
\usepackage{mathtools}
\usepackage{bm}
\usepackage{array,multirow,hhline}
\usepackage{graphicx}
\usepackage{tikz}
\usepackage[caption=false,font=footnotesize]{subfig}
\usepackage[inline]{enumitem}
\usepackage{xurl}
\usepackage[htt]{hyphenat}
\usepackage{microtype}
\usepackage{comment}
\usepackage[ruled,vlined,linesnumbered]{algorithm2e}
\usepackage[hidelinks]{hyperref}

\DontPrintSemicolon
\SetKwInput{KwIn}{Input}
\SetKwInput{KwOut}{Output}
\SetAlgoCaptionSeparator{.}

\def\BibTeX{{\rm B\kern-.05em{\sc i\kern-.025em b}\kern-.08em
T\kern-.1667em\lower.7ex\hbox{E}\kern-.125emX}}

\begin{document}

\title{Leveraging Quantum-Based Architectures for Robust Diagnostics}

\author{Shabnam~Sodagari,~\IEEEmembership{Senior Member,~IEEE,}
and Tommy~Long%
\thanks{Shabnam Sodagari and Tommy Long are with the Computer Engineering and Computer Science Department, California State University Long Beach, Long Beach, CA 90840 USA.}%
}

\maketitle

\begin{abstract}
Quantum machine learning has emerged as a promising approach for medical image analysis, particularly in settings where compact models and expressive feature representations are desired. This paper presents a hybrid classical--quantum diagnostic framework that integrates dataset-specific preprocessing, transfer learning, and quantum convolutional neural networks (QCNNs) for multi-class medical image classification. This approach is evaluated on three distinct tasks: kidney disease diagnosis from computed tomography images, cervical cell classification from pap smear images, and brain tumor classification from magnetic resonance imaging. For each dataset, a pretrained encoder is used to extract latent features, which are then embedded into quantum states through angle or amplitude encoding and processed by a QCNN. Experimental results show strong and stable convergence across all datasets. The proposed hybrid models achieve 99\% test accuracy on kidney CT classification, 97\% on cervical cell classification, and 99\% on brain tumor classification. In comparative evaluations for precision, recall, and F1, the hybrid QCNN models consistently outperform classical CNN baselines using the same pretrained encoders and similar hyperparameter settings, while requiring fewer trainable parameters. These results demonstrate the potential of quantum-enhanced architectures for robust and efficient medical diagnostics.
\end{abstract}

\section{Introduction}
\label{sec:introduction}
Recent advances in quantum machine learning have accelerated research on Quantum Convolutional Neural Networks (QCNNs) for medical image analysis. Convolutional neural networks (CNNs) have achieved strong performance in medical imaging tasks. However, given the complex features and high dimensionality present in medical imaging datasets, CNNs require large datasets and a significant amount of computational resources to achieve a high performance. CNNs also provide a limited amount of interpretability, making it difficult to distinguish the features that they have learned. The application of CNNs in the quantum computing field has shown promise in effectively addressing these challenges.

First used for classifying quantum phases, QCNNs use parameterized quantum circuits that extract features and reduce the dimensionality of quantum states \cite{cong}. QCNNs also contain fewer parameters, requiring only $O(\log N)$ parameters for $N$-qubits as compared to $O(N)$ parameters for CNNs. Several studies have extended the use of this QCNN model for classical data tasks. However, due to the limited number of qubits and significant noise present in Noisy Intermediate-Scale Quantum (NISQ) devices \cite{preskill2018}, several studies have implemented hybrid QCNN models that make use of quantum and classical layers. Hybrid QCNNs have been shown to improve model performance in medical imaging tasks by utilizing quantum entanglement and superposition to extract complex features from classical models.

 Prior research has explored QCNN-based or hybrid architectures for brain MRI classification \cite{tantawi2023brainmri}, diabetic retinopathy detection \cite{stalin2025dr}, breast cancer diagnosis \cite{xiang2024qccnn}, pneumonia classification from chest X-ray images \cite{khatoniar2024hybrid}, AI-driven healthcare analytics \cite{ovi2025multiqubit}, autism image classification using quanvolutional feature extraction \cite{cruz2025quanvolution}, and medical image classification on the Medical MNIST dataset using a quantum convolutional network combined with ResNet50 \cite{li2025medicalmnist,resnet}. These studies collectively suggest that quantum-enhanced architectures can capture complex feature relationships while maintaining compact parameterizations.

Despite these promising results, several limitations remain. Existing hybrid QCNN models often provide limited implementation details regarding architecture layout and preprocessing steps, making reproducibility difficult. Additionally, several studies focus on binary classification tasks with small-scale datasets or utilize a small number of qubits to fit the current limitations of quantum hardware. 

To address these limitations, this work combines specialized pre-processing with hybrid QCNN frameworks across multiple medical imaging datasets. In contrast to some prior studies that use limited qubit counts, fixed quanvolutional circuits, or binary settings, the present study considers multiple medical imaging tasks, i.e., computed tomography (CT) images of the kidney, pap smear images for cervical cancer, and magnetic resonance imaging (MRI) images of the brain. We leverage pretrained model encoders to extract latent representations and then process those features with a QCNN containing explicit quantum interaction layers following the architecture of \cite{mahmud2024quantum}. Within this framework, the interaction layers help capture quantum feature correlations in medical images and support hierarchical representation learning for multi-class disease classification. 

Thus, the main contributions of this paper include:
\begin{enumerate*}[label=(\roman*)]
\setlength{\leftmargin}{0pt}
    \item We adopt pretrained encoder models as the feature extraction stage of the hybrid framework, where the choice of encoder is customized for each medical dataset to better capture dataset-specific latent representations.
    \item We employ dataset-specific preprocessing strategies, including contrast-limited adaptive histogram equalization and denoising, with optimally selected parameter settings to reduce noise and enhance contrast for each medical dataset.
    \item The extracted latent features are encoded into qubits via quantum encoding such as amplitude or angle encoding and processed using a quantum CNN with interaction layers~\cite{mahmud2024quantum, xiao2025}, while we modify the last interaction layer depending on the dataset. 
    \item We perform fine-tuning on the leveraged hybrid model by unfreezing layers of the encoder to adapt to more specific features of the datasets, showing improved training performance.
    \item We evaluate the leveraged hybrid model before and after fine-tuning, demonstrating robust classification performance after fine-tuning.
    \item We compare the performance of the hybrid model to a classical CNN using the same pretrained model and hyperparameters used in the hybrid model, showcasing an increase in metrics with less trainable parameters.
\end{enumerate*}

To support reproducibility, we make our code publicly available at https://github.com/QuantumHealthResearch.
The organization of this paper is as follows. Section~\ref{sec:background} presents a review of background work on existing hybrid QCNN models. Subsequent sections describe the proposed specialized pre-processing combined with hybrid QCNN framework used for each dataset, the hyperparameters used, its training and test results, and a performance comparison with a classical CNN. Section~\ref{sec:conclus} concludes the paper.

\section{Review of Background Work}
\label{sec:background}

Several studies have extended the use of QCNNs and hybrid QCNNs to medical image analysis and related diagnostic tasks. This section reviews representative work across medical domains and non-medical imaging domains.

Among these medical imaging tasks, chest imaging has been widely used for the detection of pulmonary diseases. One study utilized a Hybrid QCNN model for detecting pneumonia from chest radiographs \cite{Baral2024}. The hybrid model was bench-marked using different types of pre-trained models as a feature extractor and was able to outperform an AlexNet pre-trained model using similar hyperparameter configurations and different dataset sizes. Another hybrid QCNN was proposed to diagnose respiratory lung diseases from chest X-ray images \cite{rao2024}. The model leveraged a custom CNN architecture, achieving near-perfect classifications with a small number of parameters and lower computing costs. Similarly, another study proposed a hybrid model for classifying lung X-ray images \cite{alfajri2026}. The model showcased high performance with a custom CNN model and specialized data augmentation. A different study focused on identifying COVID-19 from X-ray images \cite{yousif2024}. The model was able to achieve an improvement in recognition and classification accuracy compared to traditional CNNs on small datasets due to its usage of quantum superposition and parallelism. Another study utilized a custom CNN model for classifying lung cancer \cite{radhika2025}. The model utilized an optimized custom CNN model for feature extraction, achieving high accuracy and speed compared to other QCNN models. A different study focused on tuberculosis diagnosis with a custom CNN encoder. That model demonstrated linear separability in positive and negative cases. Despite the promising results, these studies focused on binary classification or achieved limited performance. In contrast, this paper focuses on applying hybrid QCNNs for multi-classification tasks and utilizing pretrained model encoders to enhance performance.

Several studies have shown promising results in classifying brain tumors with the usage of hybrid models. One study proposed a QCNN-based classification model for brain MRI images, consisting of preprocessing and QCNN-based categorization stages~\cite{tantawi2023brainmri}. The approach showed that quantum models can better capture complex spatial relationships than traditional convolutional networks. A study utilized a hybrid QCNN to classify different types of brain diseases \cite{Chandra2022}, demonstrating its capability to outperform other classical pretrained models. A different study classified brain MRI images with an adaptive hybrid QCNN \cite{ajlouni2023}. By applying quantum convolutional layers in the beginning, the model was able to outperform a classical model in both performance and size. A model was proposed for detecting the presence of dementia in brain MRI images \cite{kim2023}. By utilizing a ResNet18 pretrained model for feature extraction, the four-qubit variational quantum circuit provided an additional improvement in performance over the classical model with less trainable parameters. Similarly, a $4$-qubit model was proposed for diagnosing brain tumors \cite{kanimozhi2022}. Using a ResNet$18$ model, the model achieved higher accuracy with deeper quantum architecture depth, showcasing its ability to extract more features with the more layers added. A different study proposed a dual-branch hybrid QCNN model for brain tumor classification \cite{vijayanand2025}. The architecture combines classical convolutional mechanisms with parameterized quantum variational circuits (PQVC) to increase generalization and robustness. In contrast to prior studies, by employing a larger-qubit QCNN architecture, this paper is able to capture more informative feature representations for complex, high-dimensional, multi-class medical datasets.

Hybrid QCNNs have also had success in different medical domains. A separate investigation introduced a hybrid QCNN for diabetic retinopathy detection, integrating quantum layers with conventional convolutional ones~\cite{stalin2025dr}. The model leveraged quantum superposition and entanglement to enhance feature extraction, outperforming a purely classical CNN. Moreover, in breast cancer diagnosis, a quantum--classical hybrid convolutional neural network (QCCNN) was developed~\cite{xiang2024qccnn}. The hybrid model improved robustness and generalization compared with CNN and logistic regression models. Another study investigated QCNNs for pneumonia classification from chest X-ray images under constrained feature conditions~\cite{khatoniar2024hybrid}. The results demonstrated the robustness of QCNNs with limited data dimensionality. A multiqubit QCNN was also proposed for AI-driven healthcare analytics, utilizing parameterized quantum circuits for diagnostic tasks such as breast cancer, diabetes, and heart failure prediction~\cite{ovi2025multiqubit}. That architecture incorporated convolutional, pooling, and interaction layers to promote inter-qubit communication. Moreover, a hybrid quantum--classical framework using quanvolutional feature extraction was applied to autism image classification~\cite{cruz2025quanvolution}. The system combined a fixed four-qubit circuit with a classical classifier, showing modest accuracy gains over classical baselines. Additionally, a medical image classification model combining a quantum convolutional network with ResNet50~\cite{resnet} was applied to the Medical MNIST dataset~\cite{li2025medicalmnist}. That model demonstrated the benefits of embedding quantum operations within a classical backbone. In contrast, the QCNN used in this study uses interaction layers~\cite{mahmud2024quantum}, allowing adaptive learning of quantum feature correlations specific to medical images.

Hybrid QCNNs have also had success in different medical domains beyond imaging-intensive brain and chest tasks. A hybrid QCNN was used for skin lesion \cite{frasca2025}. By using a custom autoencoders to segment hair from skin lesions, the QCNNs were able to reduce overfitting and improve generalization that surpassed traditional CNNs. In contrast, this paper employs a pre-trained model as the encoder to extract generalized features without requiring segmentation. One study introduced a hybrid QCNN model for identifying skin cancer, utilizing quantum attention layers \cite{pandey2025}. The model incorporated attention mechanisms into the quantum architecture to reduce computational resources used and focus on relevant regions of an image. Another study focused on predicting diabetes with a sensor infused quantum CNN \cite{kotwal2025}. The model was implemented in a sensor that kept track of obesity, heart rate, hypertension, and other various health factors. The model was able to efficiently work under hardware restraints and was further processed to provide dietary recommendation, showcasing the rich features produced. A hybrid QCNN was proposed to detect stenosis in coronary angiography images \cite{li2025}. The model leveraged a Monte Carlo tree search algorithm to search for the optimal parameters for its quantum circuits, the model was able to avoid barren plateaus during training and achieve superior performance over pretrained CNN encoders.

Beyond medical applications, QCNNs have been used extensively in non-medical imaging domains. Hybrid QCNNs were used for classifying seven different types of sea animals \cite{pravin2023}. The model used a ResNet50 feature extractor and demonstrated strong generalization and robustness in all classes. A different QCNN was proposed for classifying different types of diseases found in coffee plants \cite{Kulkarni2025}. The QCNN showed rapid increases in accuracy over classical CNNs during training. Similarly, a study detected corn leaf diseases with QCNN \cite{arepalli2025}. By passing images through both classical and quantum convolutional layers in parallel, the model was able to extract complex patterns from the data and achieve near-perfect scores in accuracy and precision that outclassed a Dense CNN and GoogleNet model. A multigate quantum CNN was proposed to recognize faces \cite{zhu2024}. Using quantum convolutions, the model takes advantage of quantum entanglement to establish deeper relationships between pixels in each region of an image. This resulted in a faster running time and higher performance, while containing a smaller amount of parameters than other proposed quantum and classical models. Hybrid models were adapted to detect UAVs \cite{ahmad2025}. The architecture combined several quantum models with different tasks, showing adaptability in robustness. However, these studies focus on quantum-classical approaches where quantum layers perform the initial feature extraction and classical layers process them further. Quantum-classical models use a small number of qubits and are used for each patch of image, increasing computational costs. In this paper, we employ a classical-quantum architecture. In the following sections, we explain our dataset-specific techniques to obtain optimal results for medical classification tasks.

\section{Kidney Disease Diagnosis using Quantum-Oriented Techniques}
\label{sec:kidney}

\begin{figure}[tbh!]
    \centering
    \includegraphics[width=.5\textwidth]{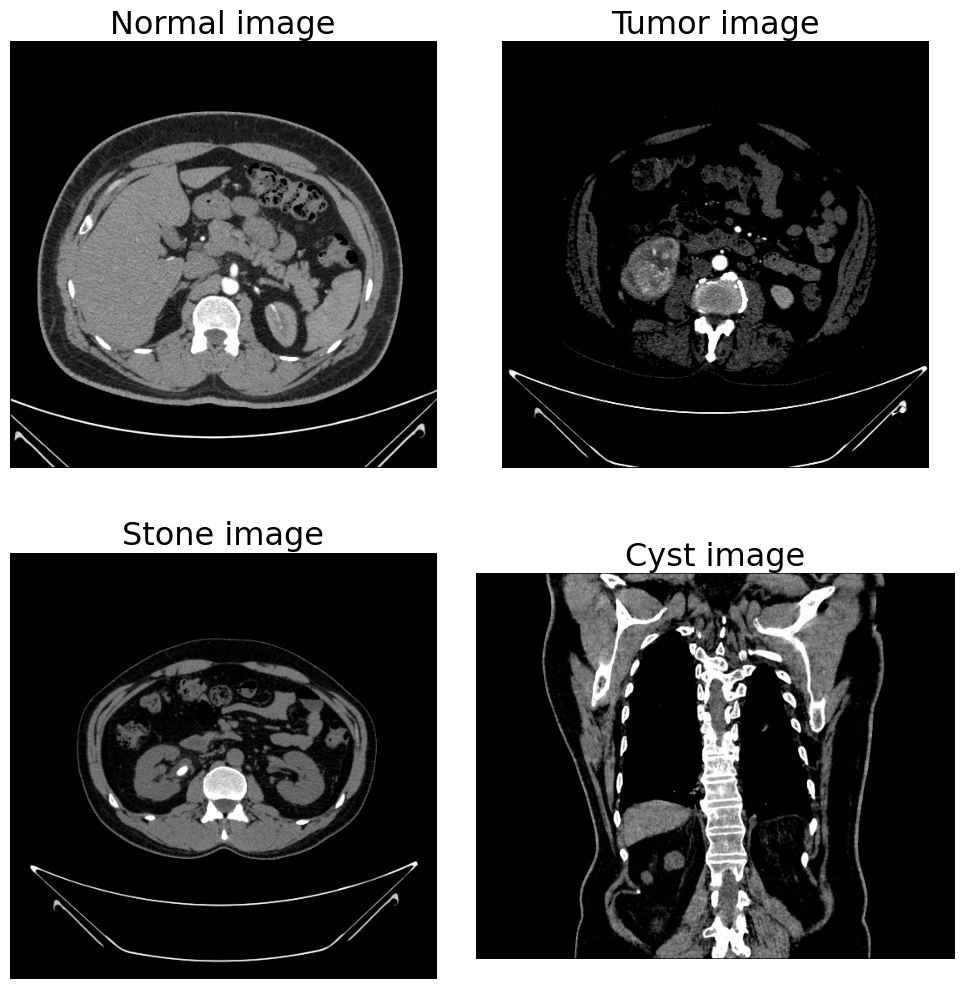}
    \caption{Sample images from kidney dataset~\cite{kaggle_kidney} displaying each of the four classes: Normal, Tumor, Stone, and Cyst.}
    \label{fig:class_imgs}
\end{figure}

The objective of this section is to diagnose and differentiate kidney stones, cysts, and tumors using CT images of the kidney. The dataset under study contains four classes: Normal with $5077$ images, Tumor with $2283$ images, Cyst with $3709$ images, and Stone with $1377$ images~\cite{kaggle_kidney}. 
Fig.~\ref{fig:class_imgs} demonstrates sample images in the dataset. 

The model that we used was built with PyTorch and PennyLane. For the preprocessing step, we split each class into their own respective training, testing, and validation sets using scikit-learn. We set the size of the training set to $80\%$, with the testing and validation set being $10\%$ each. Each class split was then instantiated as a PyTorch Dataset class object. 

\begin{figure}[tbh!]
    \centering
    \includegraphics[width=.5\textwidth]{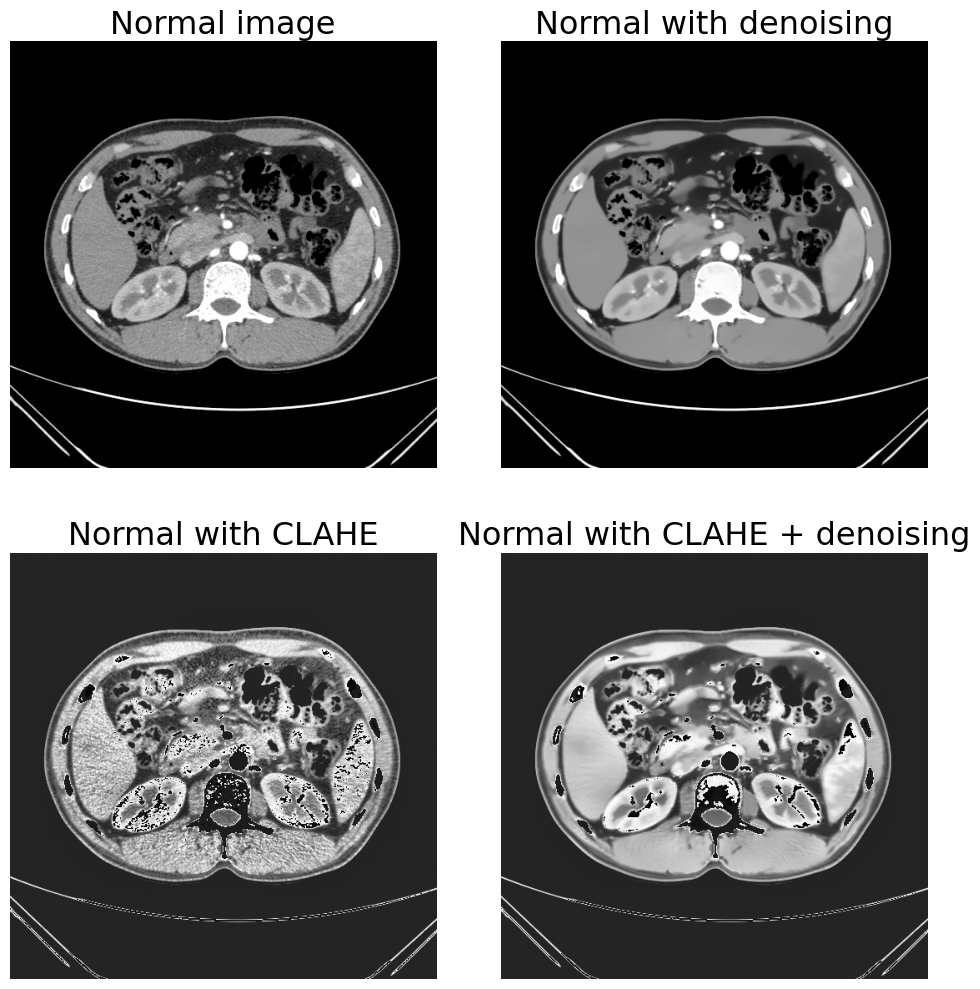}
    \caption{An example of our pre-processing of a kidney image using denoising and CLAHE.}
    \label{fig:normal_comparison}
\end{figure}

The PyTorch Dataset class object loads each image using OpenCV and converts it to grayscale. For our image pre-processing function we applied denoising and Contrast Limited Adaptive Histogram Equalization (CLAHE)~\cite{clahe}. We observed that this helps to improve feature extraction. For instance, in Fig.~\ref{fig:normal_comparison} we compare the original input image with either CLAHE or denoising or both methods applied. 

We used data augmentation to help with class imbalance. Because the Tumor and Stone classes have the least number of images, the training data for both classes were given separate data augmentation, where images would undergo various geometric transformations such as \texttt{RandomZoomOut()}, \texttt{RandomResizedCrop()}, \texttt{RandomRotation()}, and \texttt{RandomHorizontalFlip()} from the PyTorch \texttt{transforms} library to increase dataset variety.

Algorithm~\ref{alg:hybrid_qcnn_kidneyct} summarizes the steps. 
\begin{algorithm}[!t]
\caption{Kidney CT Classification (12 qubits: 8 data $q_0\!:\!q_7$ + 4 ancilla $a_0\!:\!a_3$)}
\label{alg:hybrid_qcnn_kidneyct}
\footnotesize
\DontPrintSemicolon
\KwIn{Image $x$; class set $\{0,1,2,3\}$}
\KwOut{$\hat{\mathbf{y}}\in\mathbb{R}^4$}
\BlankLine

\textbf{Denoise \& Enhance:}
grayscale $\rightarrow$ fast NLM $(10,7,21)$ $\rightarrow$ CLAHE (clip$=5$) $+30$ shift $\rightarrow$ clip to $[0,255]$.\;

\textbf{Encode \& Embed:} ResNet50:
$\mathbf{z}\!\leftarrow\!\mathrm{Enc}(x)\in\mathbb{R}^8$; rescale to $[0,\pi]^8$;\;
Apply $R_Y(z_j)$ on each $q_j$.\;

\textbf{Convolution Layer 1 (2 passes; $15$ params each:}
apply the convolutional circuit~9 of \cite{Hur2022} on
$(q_0,q_7)$, then $(q_0,q_1),(q_2,q_3),(q_4,q_5),(q_6,q_7)$, then $(q_1,q_2),(q_3,q_4),(q_5,q_6)$;
repeat with new $15$ parameters.\;

\textbf{Pooling (2 params; \cite{Hur2022}):}
for $i\!\in\!\{0,2,4,6\}$ apply $\mathrm{CRZ}(\phi_0)$–$X$–$\mathrm{CRX}(\phi_1)$ on $(q_{i+1}\!\rightarrow q_i)$;
retain $q_0,q_2,q_4,q_6$.\;

\textbf{Interaction Layer 1 (paramless; \cite{mahmud2024quantum}):}
Toffoli cascade among $\{q_0,q_2,q_4,q_6\}$ to establish entanglement.\;

\textbf{Convolution Layer 2 (2 passes; $15$ params each):}
on the 4 active data wires (reindexed locally $0..3$), apply the same ansatz on
$(0,3)$, $(0,1)$, $(2,3)$, $(1,2)$; repeat with new parameters
(\emph{equivalently}: physical pairs $(0,6)$, $(0,2)$, $(4,6)$, $(2,4)$).\;

\textbf{Interaction Layer 2 ($12$ params; \cite{mahmud2024quantum}):}
on $\{q_0,q_2,q_4,q_6\}$ apply Toffoli–$R_X$–Toffoli–$R_Y$–Toffoli–$R_Z$.\;

\textbf{Classifier Interaction ($12$ params:}
CNOT ring on data $q_6{\rightarrow}q_0{\rightarrow}q_2{\rightarrow}q_4{\rightarrow}q_6$;
for $k{=}0..3$: $\mathrm{CNOT}(q_{2k}{\rightarrow}a_k)$ then $R_Z(\beta_{3k})\,R_Y(\beta_{3k+1})\,R_Z(\beta_{3k+2})$ on $a_k$\;
Readout $\mathbf{q}=(\langle Z\rangle_{a_0},\langle Z\rangle_{a_1},\langle Z\rangle_{a_2},\langle Z\rangle_{a_3})$.\;

\textbf{Classical Head:} MLP$(\mathbf{q})\!\to\!\hat{\mathbf{y}}$.\;
\end{algorithm}

More specifically, we passed our datasets into a PyTorch DataLoader where \texttt{WeightedRandomSampler()} was used for the training dataset to help with sampling minority classes such as Tumor and Stone. The sample size parameter in \texttt{WeightedRandomSampler()} was determined by multiplying the number of classes, $4$, with an adjustable number of samples per class, $n_{\text{max}}$, to form $n_{\text{max}} \times 4$ samples drawn per epoch with replacement.

Other class data like Normal and Cyst were given a basic transform where we only applied grayscaling.

We then concatenated all separate class dataset splits together to form one train, test, and validation dataset. 

At the end of both transforms, we applied a ResNet50 preprocessing function. 
This helped prepare the images for input to an encoder part of an autoencoder where the ResNet50 pretrained model was used for feature extractions and inference of latent variables. For feature extraction we froze the ResNet50 model's weights during training, except that we adjusted the final layer of the pretrained model to have two fully connected \texttt{Linear} layers that reduced from $2048$ neurons (to ReLU) to $256$ and then to a tunable number of latent variables. Although the original implementation uses an autoencoder that reconstructs the image with a single \texttt{Linear} layer \cite{mahmud2024quantum}, we expand upon it by utilizing a pre-trained model due to the dataset containing a higher level of dimensionality.

We fed the latent variables that we obtained at the output of our ResNet50 encoder to an angle encoder to represent classical variables as qubits. In angle encoding each qubit is rotated using the $R_{y}$ gate. The latent variables are then rescaled to the range of $[0, \pi]$ before being fed to the angle encoder. The output of the angle encoding scheme was the input to a particular QCNN architecture with interaction~\cite{mahmud2024quantum} layers.

More specifically, after angle encoding, the qubits are passed into an ansatz composed of an architecture with $15$ parameters utilizing $U_3(\theta, \phi, \lambda) = R_z(\phi) \, R_x(-\tfrac{\pi}{2}) \, R_z(\theta) \, R_x(\tfrac{\pi}{2}) \, R_z(\lambda)$, as well as $R_y$ and $R_z$ rotation gates~\cite{mahmud2024quantum}. The ansatz is used as the quantum convolutional layer for the hybrid QCNN model. The ansatz is applied to each even-numbered qubit and its neighboring qubit. For example, if the input was $8$ qubits, the physical pairs would be: ($(0, 7)$, $(0, 1)$, $(2, 3)$, $(4, 5)$, $(6, 7)$). The same is repeated for the odd-numbered qubits and its neighboring qubits, visualized as physical pairs ($(1, 2)$ $(3, 4)$, $(5, 6)$). Applying the ansatz to each pairing establishes quantum entanglement. This convolutional layer is repeated two times and the output is fed into a quantum pooling layer with parameterized $Z$ and $X$ rotation gates, where the number of qubits is reduced by half, keeping only the even-numbered qubits. After pooling, the remaining qubits go into the first interaction layer~\cite{mahmud2024quantum}, which uses Toffoli gates to establish three-qubit interactions and create entanglement between the qubits.  

Following the qcnn of~\cite{mahmud2024quantum}, a second quantum convolutional layer is then applied, the output of which is fed into the second interaction layer. In Convolutional Layer~2 (two passes), the same two-qubit ansatz used in the first convolutional layer is employed, maintaining $15$ trainable parameters per pass. After pooling reduces the number of active data qubits to $\{q_0,q_2,q_4,q_6\}$, the ansatz is applied to a new connectivity pattern among these qubits—specifically, on local pairs $(0,3)$, $(0,1)$, $(2,3)$, and $(1,2)$ (corresponding to physical pairs $(0,6)$, $(0,2)$, $(4,6)$, and $(2,4)$).
The second interaction layer with $12$ parameters is similar to the first interaction layer, but uses $R_{x}$, $R_{y}$, and $R_{z}$ rotation gates along with trainable parameters between Toffoli gates~\cite{mahmud2024quantum}. The last interaction layer takes the remaining qubits and entangles these qubits with the ancilla qubits. The ancilla qubits represent the number of classes which is $4$ for the kidney CT images under study. The $4$ ancilla qubits are then passed into a classical network with $3$ \texttt{Linear} layers with \texttt{ReLU} activations in between each \texttt{Linear} layer.  
In addition to addressing class imbalance with targeted data augmentation, we compute the class weights to be used for the CrossEntropyLoss function with the following equation: $w_c = \frac{1}{max(n_c, 1)}$, $n_c$ being the number of samples in class $c$ and $w_c$ being the weight for class $c$. This gives classes with less samples more weight while classes with more samples receive less weight. 

We tested the method on the CT kidney dataset~\cite{kaggle_kidney} using various hyperparameters, such as learning rate, batch size, sample size, number of qubits, etc. 
We used the Adam optimizer and a class-weighted cross-entropy loss. Each training epoch drew $4,000$ samples, $1,000$ per class across the four diagnostic categories (normal, cyst, stone, tumor), using a weighted random sampler to maintain class balance despite dataset size differences. Sampling with replacement ensured equal representation of underrepresented classes, while the choice of $1,000$ samples per class balanced computational efficiency with statistical representativeness.
We considered latent dimensions of $8$ and $4$, which, combined with $4$ ancilla qubits for class encoding, yield $12$-qubit and $8$-qubit systems, respectively.

\subsection{Results and Analysis of Kidney CT Images}

As shown in Table~\ref{tab:qubit_metrics}, the $8$-qubit model shows stronger performance on several key metrics, particularly in Normal and Stone classification. It achieves perfect precision for Cyst and perfect recall for Stone. It also delivers higher micro-precision and micro-recall compared to the $12$-qubit model.

\begin{table}[htbp]
\caption{Performance metrics comparison between 8-qubit and 12-qubit models for kidney disease classification.}
\label{tab:qubit_metrics}
\centering
\footnotesize
\setlength{\tabcolsep}{3pt}
\renewcommand{\arraystretch}{0.92}
\begin{tabular}{@{}llcc@{}}
\hline
\textbf{Class} & \textbf{Metric} & \textbf{8Q} & \textbf{12Q} \\
\hline
\multirow{3}{*}{Normal}
  & Precision & 0.9941 & 0.9941 \\
  & Recall    & 0.9980 & 1.0000 \\
  & F1 Score  & 0.9961 & 0.9971 \\
\hline
\multirow{3}{*}{Cyst}
  & Precision & 1.0000 & 0.9867 \\
  & Recall    & 0.9919 & 1.0000 \\
  & F1 Score  & 0.9959 & 0.9933 \\
\hline
\multirow{3}{*}{Stone}
  & Precision & 0.9787 & 1.0000 \\
  & Recall    & 1.0000 & 0.9855 \\
  & F1 Score  & 0.9892 & 0.9927 \\
\hline
\multirow{3}{*}{Tumor}
  & Precision & 0.9956 & 1.0000 \\
  & Recall    & 0.9869 & 0.9738 \\
  & F1 Score  & 0.9912 & 0.9867 \\
\hline
Macro & Precision & 0.9921 & 0.9952 \\
Micro & Precision & 0.9944 & 0.9936 \\
Macro & Recall    & 0.9942 & 0.9898 \\
Micro & Recall    & 0.9944 & 0.9936 \\
Macro & F1        & 0.9931 & 0.9924 \\
Micro & F1        & 0.9944 & 0.9936 \\
MCC   &           & 0.9920 & 0.9908 \\
Test Accuracy &   & 0.99   & 0.99   \\
\hline
\end{tabular}
\vspace{-0.4em}
\end{table}

\begin{figure*}[!tbp]
\centering
\subfloat[8 qubits]{%
    \includegraphics[width=0.48\textwidth]{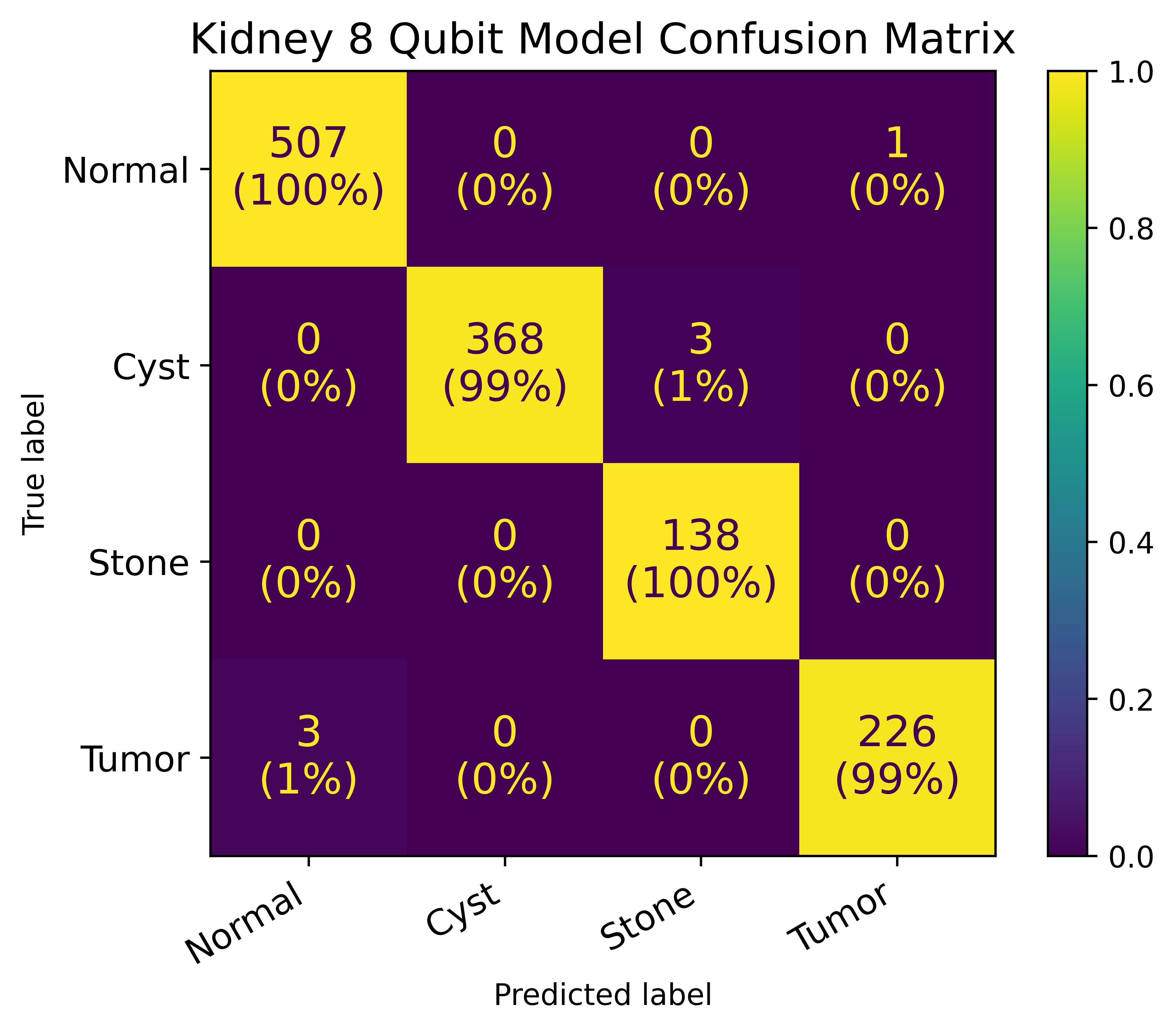}%
    \label{fig:kidney_cm_8q}}
\hfill
\subfloat[12 qubits]{%
    \includegraphics[width=0.48\textwidth]{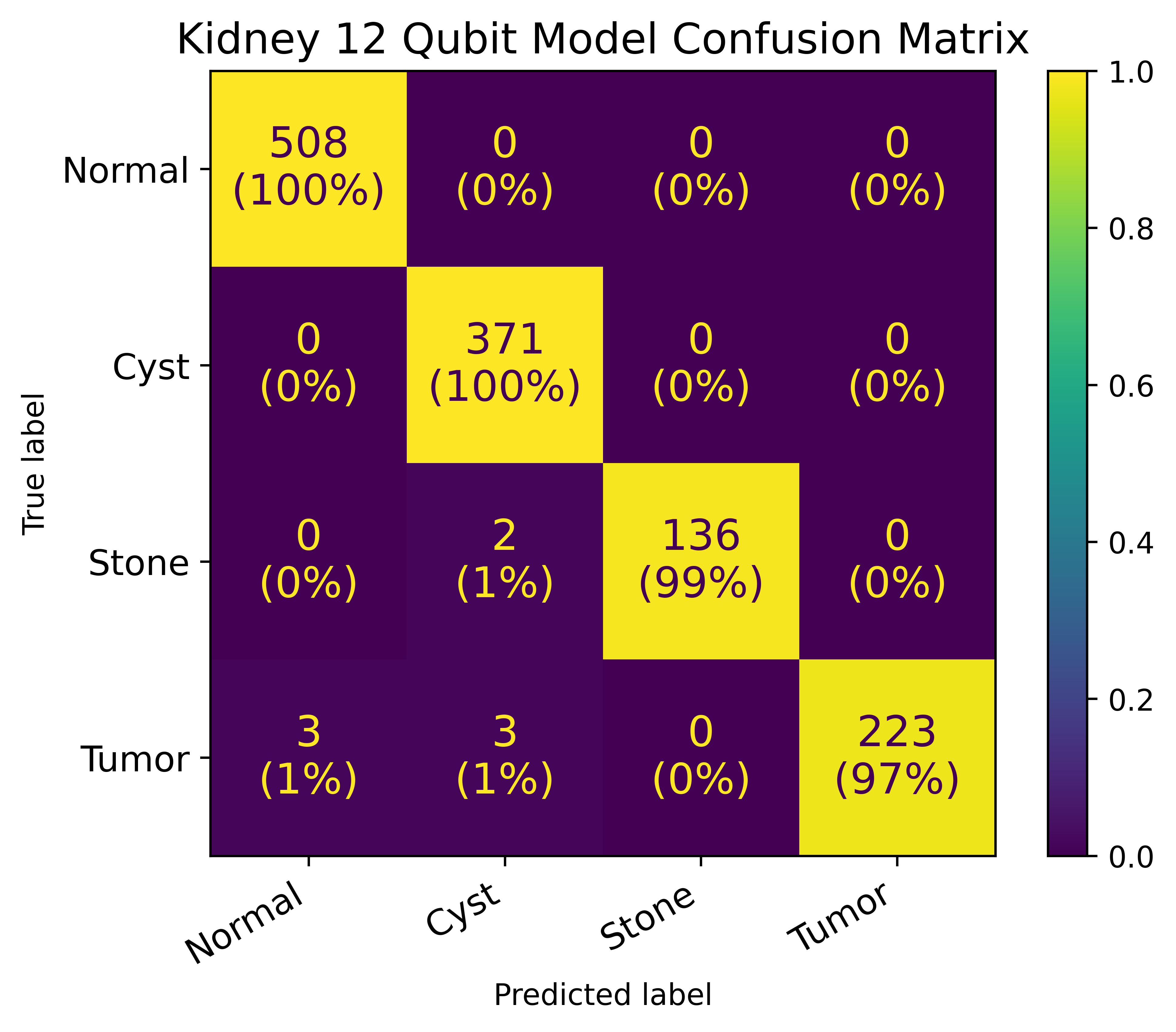}%
    \label{fig:kidney_cm_12q}}
\caption{Comparison of kidney confusion matrices for (a) 8 qubits and (b) 12 qubits, both trained with a learning rate of $0.001$, batch size of $8$, and sample size of $4000$.}
\label{fig:conf_matrices}
\end{figure*}

Figure~\ref{fig:conf_matrices} compares the confusion matrices for $8$ qubits and $12$ qubits both with a learning rate of $0.001$, batch size of $8$, and sample size of $4000$.

\begin{figure*}[!tbp]
\centering
\subfloat[8 qubits]{%
    \includegraphics[width=0.48\textwidth]{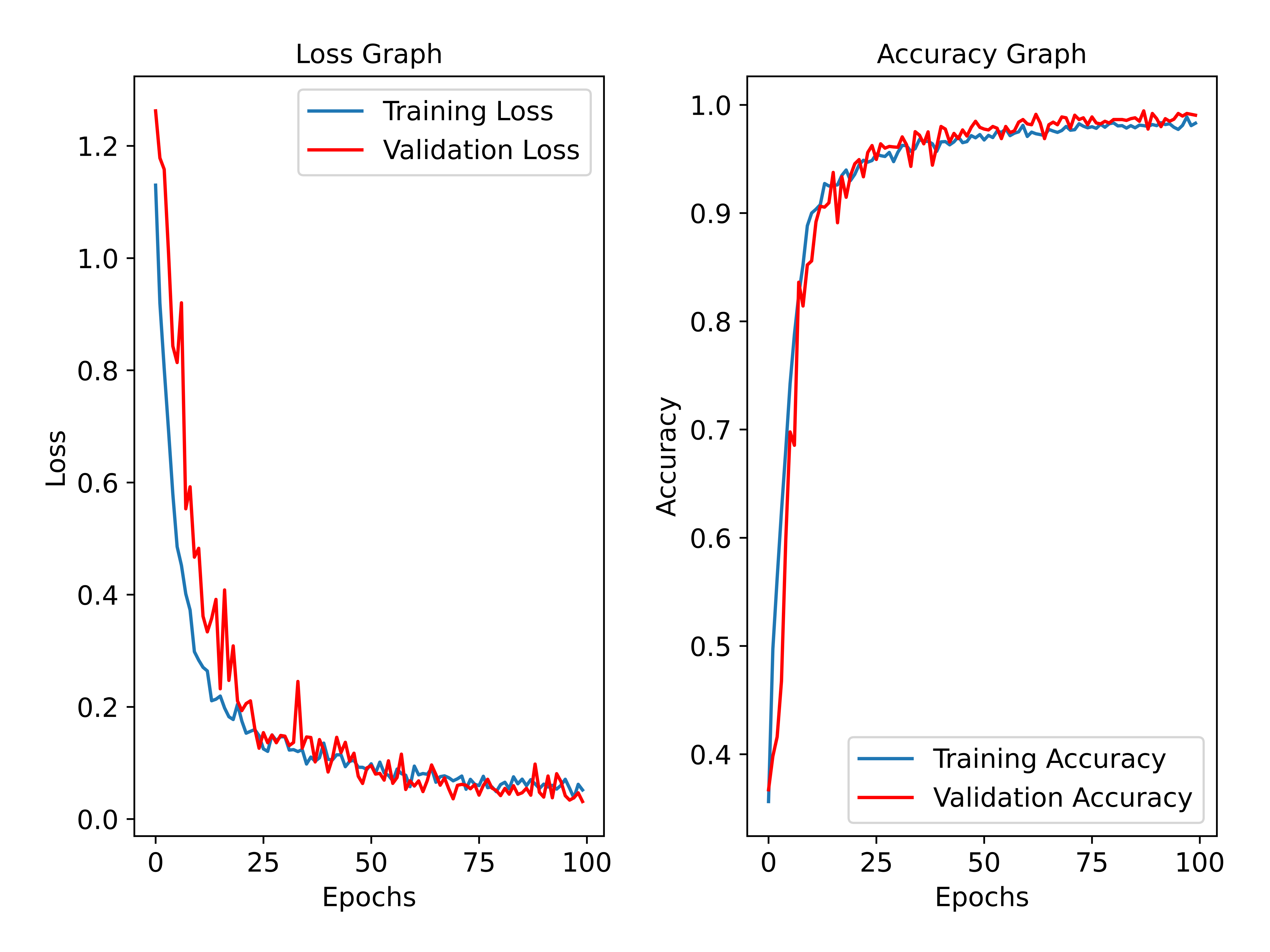}%
    \label{fig:8qbt}}
\hfill
\subfloat[12 qubits]{%
    \includegraphics[width=0.48\textwidth]{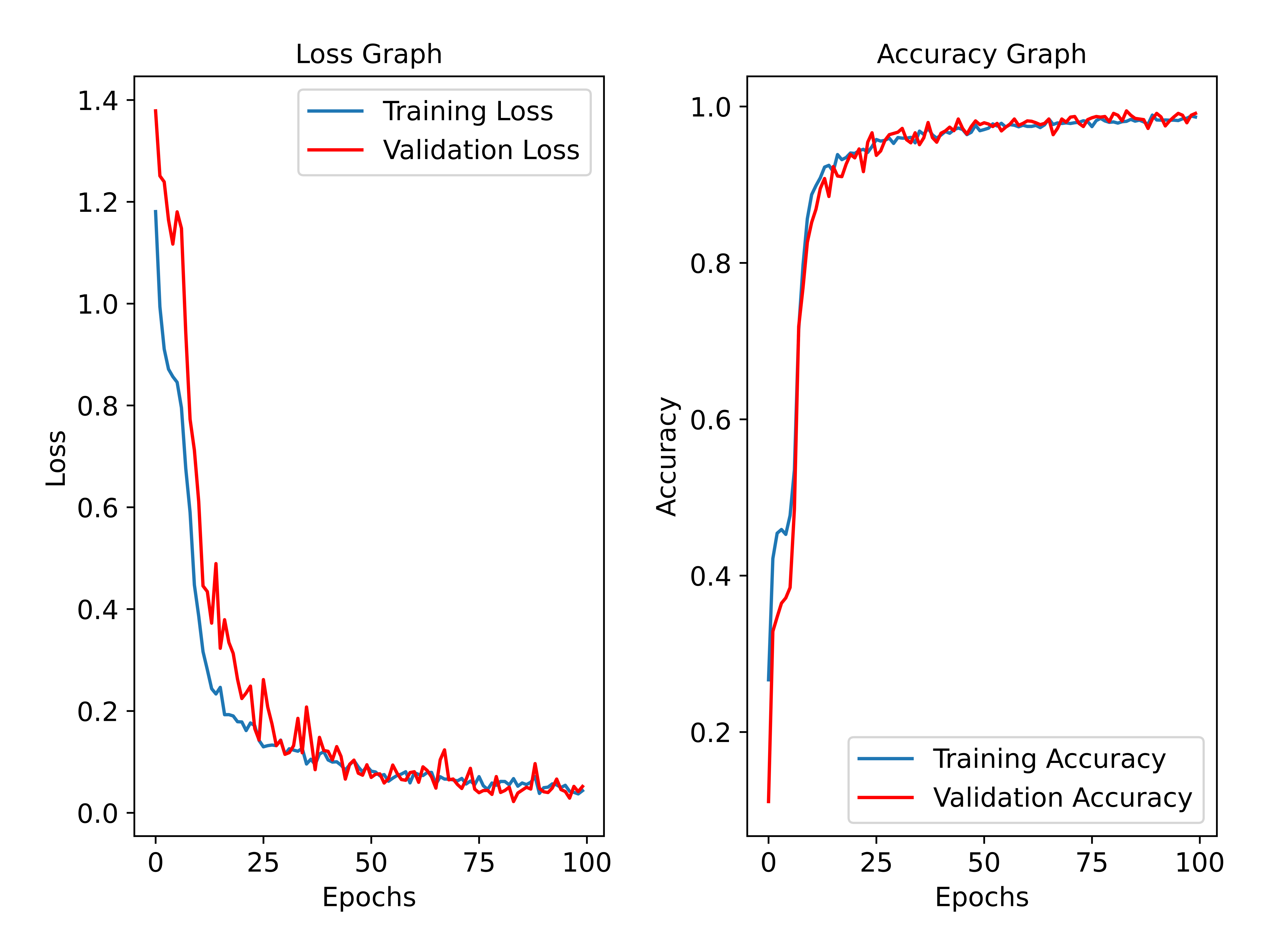}%
    \label{fig:12qbt}}
\caption{Comparison of kidney training and validation accuracy and loss for (a) $8$ qubits and (b) $12$ qubits, both trained over $100$ epochs with a learning rate of $0.001$, batch size of $8$, and sample size of $4000$.}
\label{fig:results}
\end{figure*}

Figure~\ref{fig:results} compares the training and validation accuracy and loss for the hyperparameters in Fig.~\ref{fig:conf_matrices}.

As illustrated in Figure~\ref{fig:conf_matrices}, both models show excellent accuracy for the Normal, Cyst, and Stone classes. However, minor performance differences are observed in the Tumor category, with the $8$-qubit model classifying more correct images than the $12$-qubit model. The $8$-qubit model achieves slightly better separation, with fewer instances of Stone and Tumor cases being misclassified. While both configurations perform well, the $8$-qubit model demonstrates higher robustness, particularly in clinically sensitive tumor differentiation.

As shown in Figure~\ref{fig:results} the loss and accuracy curves for both models demonstrate stable convergence and strong generalization. This indicates effective learning with no clear signs of overfitting. The validation accuracy closely tracks the training accuracy throughout. The $8$-qubit model achieves marginally lower final loss and slightly more stable validation accuracy near convergence.

Next, we address the cervical cell classification through our proposed dataset-specific pre-processing while leveraging hybrid quantum architectures.

\section{Cervical Cell Classification in Pap Smear Images}
\label{sec:cervic}

\begin{figure*}[!tbp]
    \centering
    \includegraphics[width=\textwidth]{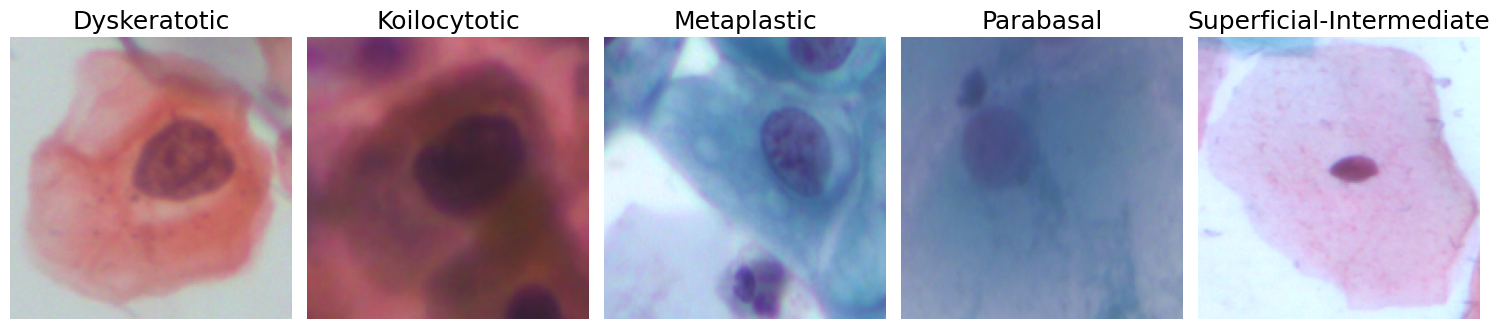}
    \caption{Sample images from cervical cancer dataset~\cite{plissiti2018} displaying each of the five classes: Koilocytotic, Metaplastic, Parabasal, and Superficial-Intermediate.}
    \label{fig:cervical_class_imgs}
\end{figure*}

We evaluate the performance of hybrid QCNNs on classfying different types of cervical cancer cells in pap smear images \cite{plissiti2018}. The dataset under study contains $5$ classes: Dyskeratotic with $813$ images, Koilocytotic with $825$ images, Metaplastic with $793$ images, Parabasal with $787$ images, and Superficial-Intermediate with $831$ images. An image of each class is shown in Figure~\ref{fig:cervical_class_imgs}. We set the size of the training set to $80\%$ of the dataset, with the testing and validation set being $10\%$ each. The labels were stratified to ensure that there was an equal amount of samples per class in each set. 

\begin{figure}[tbh!]
    \centering
    \includegraphics[width=.5\textwidth]{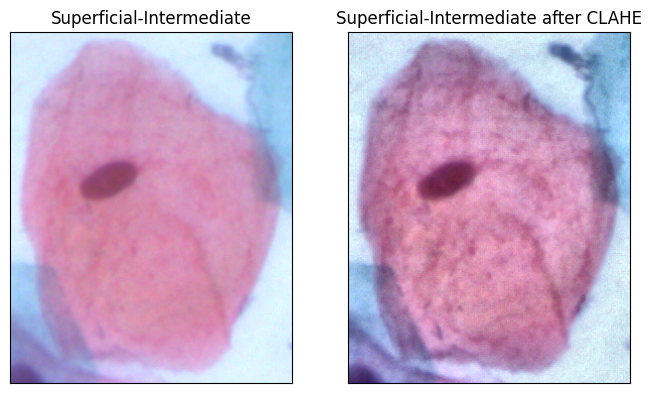}
    \caption{A comparison of an original image of the Superficial-Intermediate cervical cancer cell vs. when applying CLAHE.}
    \label{fig:cervical_clahe}
\end{figure}

Each image retrieved from the dataset was loaded using OpenCV and converted to RGB color. To improve the image's low contrast and lighting, the image is converted from RGB to a new color space composed of three different components: lightness (L), green to magenta (A), and blue to yellow (B) which makes up the LAB color space. We use LAB because this helps to separate the brightness information from the colors itself. We apply CLAHE \cite{clahe} on the lightness component of the image and merge the result with the other color components to form a new image with better lighting and contrast. The new image is shown in comparison with the original image in Figure~\ref{fig:cervical_clahe}. By improving the contrast, the new image shows darker outlines around the cell, highlighting more features. To improve generalization, data augmentation was used for the training dataset. The images underwent actions such as cropping, vertical and horizontal flipping, rotations, and an EfficientNetB4 transformation function. 

\begin{figure*}[tbp!]
    \centering
    \resizebox{\textwidth}{!}{%
        \usetikzlibrary{positioning,fit,arrows.meta,calc,backgrounds}

\definecolor{softblue}{RGB}{220,232,243}
\definecolor{softteal}{RGB}{225,239,236}
\definecolor{softmint}{RGB}{230,239,230}
\definecolor{softlavender}{RGB}{235,231,244}
\definecolor{softpeach}{RGB}{246,235,227}
\definecolor{softrose}{RGB}{246,234,234}
\definecolor{softsand}{RGB}{242,238,226}
\definecolor{softlilac}{RGB}{239,235,244}
\definecolor{linegray}{RGB}{70,76,84}
\definecolor{dashgray}{RGB}{110,118,128}

\begin{tikzpicture}[
  font=\sffamily,
  >=Latex,
  node distance=8mm and 12mm,
  flow/.style={-Latex, line width=1.05pt, draw=linegray},
  lineonly/.style={line width=1.05pt, draw=linegray},
  block/.style={
    draw=linegray,
    rounded corners=2.8pt,
    line width=0.95pt,
    minimum height=9mm,
    minimum width=68mm,
    align=center,
    inner sep=4pt,
    fill=white
  },
  aux/.style={
    draw=linegray,
    rounded corners=2.8pt,
    line width=0.9pt,
    minimum height=8mm,
    minimum width=35mm,
    align=center,
    inner sep=3pt,
    fill=softsand
  },
  outlabel/.style={
    fill=white,
    inner xsep=2.3pt,
    inner ysep=1.2pt,
    font=\normalsize,
    text=black
  }
]

\node[font=\bfseries\large, text=black] (input) {Cervical Images};
\node[block, fill=softblue, below=of input] (prep) {Preprocessing (CLAHE)};
\node[block, fill=softteal, below=of prep] (enc) {EfficientNetB4 Encoder};
\node[block, fill=softmint, below=of enc] (fc) {Fully Connected Layer};
\node[block, fill=softpeach, below=of fc] (embed) {Amplitude Embedding};
\node[
  block,
  fill=softlavender,
  minimum height=14mm,
  below=16mm of embed
] (hybrid) {%
  \textbf{Hybrid QCNN with interaction layers}~\protect\cite{mahmud2024quantum, xiao2025}%
};
\node[block, fill=softlilac, below=16mm of hybrid] (rot) {Rotation gates\\Quantum measurements};
\node[block, fill=softrose, below=of rot] (prob) {Probabilities};
\node[block, fill=softsand, below=of prob] (loss) {Cross-Entropy Loss};

\begin{scope}[on background layer]
  \node[
    draw=dashgray,
    dashed,
    rounded corners=4pt,
    line width=1pt,
    fit=(enc)(hybrid),
    inner sep=8pt
  ] (model) {};
\end{scope}

\draw[flow] (input) -- (prep);
\draw[flow] (prep) -- node[outlabel,right] {384 $\times$ 384 image} (enc);
\draw[flow] (enc) -- node[outlabel,right] {1792 features} (fc);
\draw[flow] (fc) -- node[outlabel,right] {256 features} (embed);
\draw[flow] (embed) -- (hybrid);
\draw[flow] (hybrid) -- (rot);
\draw[flow] (rot) -- (prob);
\draw[flow] (prob) -- (loss);

\node[aux, left=31mm of enc] (unfreeze) {Unfreeze 2 layers};
\node[font=\normalsize, text=black, anchor=east] (fttxt) at ([xshift=-7mm]unfreeze.west) {Fine-tune};
\coordinate (ftbase) at ([yshift=-10mm]unfreeze.south);
\coordinate (ftleft) at ([xshift=-34mm]ftbase);
\draw[flow] (fttxt.east) -- (unfreeze.west);
\draw[lineonly] (loss.west) -| (ftleft);
\draw[lineonly] (ftleft) -- (ftbase);
\draw[flow] (ftbase) -- (unfreeze.south);
\draw[flow] (unfreeze.east) -- (enc.west);

\node[aux, fill=softmint, left=31mm of hybrid, yshift=-17mm] (update) {Update parameters};
\draw[flow] (loss.west) -| (update.south);
\draw[flow] (update.north) |- ([xshift=-9mm]model.west) -- (model.west);

\end{tikzpicture}
    }
    \caption{Block diagram of hybrid classical-quantum 11-qubit model for cervical cancer diagnosis. The images go through classical feature extraction with EfficientNetB4. Latent features converted into qubits by amplitude embedding are further processed by quantum convolutional and interaction layers to establish quantum entanglement and extract deeper features for fine tuning.}
    \label{fig:cervicalhqcnn}
\end{figure*}

We use amplitude encoding as the quantum encoder due to its ability to encode features of size $2^n$, $n$ being the number of qubits that represent those features. Whether amplitude encoding performs better than angle embedding can vary depending on the dataset. The $256$ feature output from the encoder is converted into $8$ qubits and fed into the QCNN model.

\begin{figure}[htbp!]
    \centering
\includegraphics[width=\linewidth,height=0.42\textheight,keepaspectratio]{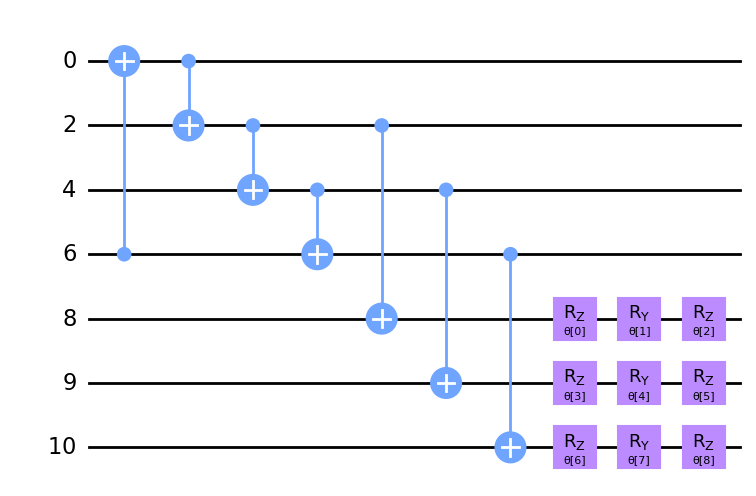}
    \caption{The third interaction layer for cervical diagnosis. Unlike the third interaction layer in~\cite{mahmud2024quantum}, which assigns one ancilla qubit to each class, the proposed layer uses the joint superposition of three ancilla qubits to encode eight probabilistic class states~\cite{xiao2025}. Three of these states are discarded~\cite{mordacci2024}, leaving five states for the five-class cervical dataset.}
    \label{fig:cc_third_interact}
\end{figure}

A block diagram of the hybrid model for cervical cancer from pap smear data is shown in Figure~\ref{fig:cervicalhqcnn}. 
The 8 qubits are fed into the first quantum convolutional layer. This layer uses the same parameterized quantum circuit (PQC) as in Fig. 3 of \cite{xiao2025}. The quantum pooling layer, quantum convolutional layer 2, interaction layer
1, and interaction layer 2 are similar to \cite{mahmud2024quantum}; however, the third interaction layer is modified to include 3 ancilla qubits, as shown in Fig.~\ref{fig:cc_third_interact}.

Rather than having ancilla qubits represent the number of classes to classify, we measure the probability values for the ancilla register~\cite{xiao2025, mordacci2024} to get a total of \(2^n\) possible quantum states. This enables the model to classify datasets containing more than $3$ classes. Because the cervical cancer cell dataset contains only $5$ classes, we use only $5$ out of the $8$ quantum states and discard the remaining. The probability values are converted into classical logit values that match the expected value range of the cross entropy loss function. To improve training stability, the QCNN parameters were initialized from a uniform distribution over $[-1,1]$. We additionally applied gradient clipping with a maximum norm of $1.0$ during optimization. 

Algorithm~\ref{alg:hybrid_qcnn_cervical_ft} summarizes the steps. 

\begin{algorithm*}[tbh!]
\caption{Cervical Cancer Classification (11 qubits: 8 data $q_0\!:\!q_7$ + 3 ancilla: $a_0\!:\!a_2$)}
\label{alg:hybrid_qcnn_cervical_ft}
\footnotesize
\DontPrintSemicolon
\KwIn{Train/valid image sets; class set $\{0,1,2,3,4\}$}
\KwOut{Best fine-tuned hybrid model; $\hat{\mathbf{y}}\in\mathbb{R}^5$}
\BlankLine

\textbf{Enhance:}
RGB $\rightarrow$ LAB $\rightarrow$ CLAHE (clip$=2$, tile$=8\times 8$) on L-color component $\rightarrow$ merge(L, a, b) $\rightarrow$ RGB.\;

\textbf{Transfer learning setup:}
load pretrained EfficientNetB4; freeze all backbone parameters; replace classifier by $\mathrm{Linear}(1792,256)$.\;

\textbf{Build hybrid model:}
use 8 data qubits $q_0\!:\!q_7$ and 3 ancillas $a_0\!:\!a_2$; load the best pre-fine-tuning hybrid checkpoint.\;

\textbf{Encode \& Embed:} EfficientNetB4:
$\mathbf{z}\!\leftarrow\!\mathrm{Enc}(x)\in\mathbb{R}^{256}$.\;

\textbf{Apply Amplitude Embedding:}
$\mathbb{R}^{256}$ $\rightarrow$ $\log_2(256)$ $\rightarrow$ 8 qubits.\;

\textbf{Convolution Layer 1 (2 passes; $15$ params each):}
apply the convolutional circuit of \cite{xiao2025} on
$(q_0,q_7)$, then $(q_0,q_1),(q_2,q_3),(q_4,q_5),(q_6,q_7)$, then $(q_1,q_2),(q_3,q_4),(q_5,q_6)$;
repeat with same $15$ parameters.\;

\textbf{Pooling (2 params; \cite{Hur2022}):}
for $i\!\in\!\{0,2,4,6\}$ apply $\mathrm{CRZ}(\phi_0)$--$X$--$\mathrm{CRX}(\phi_1)$ on $(q_{i+1}\!\rightarrow q_i)$;
retain $q_0,q_2,q_4,q_6$.\;

\textbf{Interaction Layer 1 (paramless; \cite{mahmud2024quantum}):}
apply the Toffoli cascade on $\{q_0,q_2,q_4,q_6\}$.\;

\textbf{Convolution Layer 2 (2 passes; $15$ params each):}
on the 4 active data wires, apply the same ansatz on
$(q_0,q_6)$, $(q_0,q_2)$, $(q_4,q_6)$, $(q_2,q_4)$;
repeat with same parameters.\;

\textbf{Interaction Layer 2 ($12$ params; \cite{mahmud2024quantum}):}
on $\{q_0,q_2,q_4,q_6\}$ apply
Toffoli--Toffoli--$R_X$ on all 4 wires--Toffoli--$R_Y$ on all 4 wires--Toffoli--$R_Z$ on all 4 wires.\;

\textbf{Classifier Interaction ($9$ params):}
CNOT ring on data $q_6{\rightarrow}q_0{\rightarrow}q_2{\rightarrow}q_4{\rightarrow}q_6$;
apply $\mathrm{CNOT}(q_2{\rightarrow}a_0)$, $\mathrm{CNOT}(q_4{\rightarrow}a_1)$, and
$\mathrm{CNOT}(q_6{\rightarrow}a_2)$;
then apply $\mathrm{Rot}(\beta_0,\beta_1,\beta_2)$ on $a_0$,
$\mathrm{Rot}(\beta_3,\beta_4,\beta_5)$ on $a_1$,
and $\mathrm{Rot}(\beta_6,\beta_7,\beta_8)$ on $a_2$.\;

\textbf{Measure ancillas~\cite{xiao2025}:}
$\mathbf{p}^{(8)}=(p_{000},p_{001},p_{010},p_{011},p_{100},p_{101},p_{110},p_{111})$.\;

\textbf{Keep 5 classes:}
retain the first five states, $\mathbf{p}^{(5)}=(p_{000},p_{001},p_{010},p_{011},p_{100})$.\;

\textbf{Logits:}
normalize as implemented in the model,
$\tilde{\mathbf p}=\mathbf p^{(5)}\big/\max\!\left(\sum_j p_j^{(5)},10^{-10}\right)$,
then compute $\hat{\mathbf y}=\log(\tilde{\mathbf p})$.\;

\textbf{Fine-tuning search:}
from the loaded pre-fine-tuning checkpoint, read batch size and base learning rate $\eta$;
for each Optuna trial, choose blocks
$N\!\in\!\{1,2,3,4\}$,
$\eta_e\!\in\![\eta/10,\eta/5]$,
and $\eta_f\!\in\![\eta,5\eta]$.\;

\textbf{Unfreeze \& optimize:}
unfreeze only the top $N$ EfficientNet feature blocks;
keep earlier blocks frozen;
use Adam with learning rate $\eta_e$ for unfrozen encoder blocks, and $\eta_f$ for the encoder classifier and QCNN.\;

\textbf{Train:}
minimize cross-entropy on $\hat{\mathbf y}$;
keep BatchNorm layers in frozen blocks in eval mode;
clip gradients to norm $1.0$.\;

\textbf{Select model:}
save the checkpoint with the best validation macro recall; return the best fine-tuned model.\;
\end{algorithm*}

The hybrid QCNN model was trained for a total of $200$ epochs. Because the pretrained model of the encoder is frozen, we set its batch normalization layers to evaluation mode in PyTorch after each epoch to prevent the mean and variance from updating during training. This kept its mean and variance as the original parameters it was trained on. The Adam optimizer was used to update the model's parameters and cross entropy was used for the loss function. The Optuna library was used to find the optimal hyperparameters for batch size and learning rate during training by testing out different values through multiple trials \cite{akiba2019}. We define our search space for learning rate between $0.1$ and $0.00001$ while batch size was between $8$ and $64$.

\subsection{Results and Analysis on Pap Smear Images of Cervical Cells}

\begin{figure}[tbh!]
        \centering
        \includegraphics[width=\linewidth]{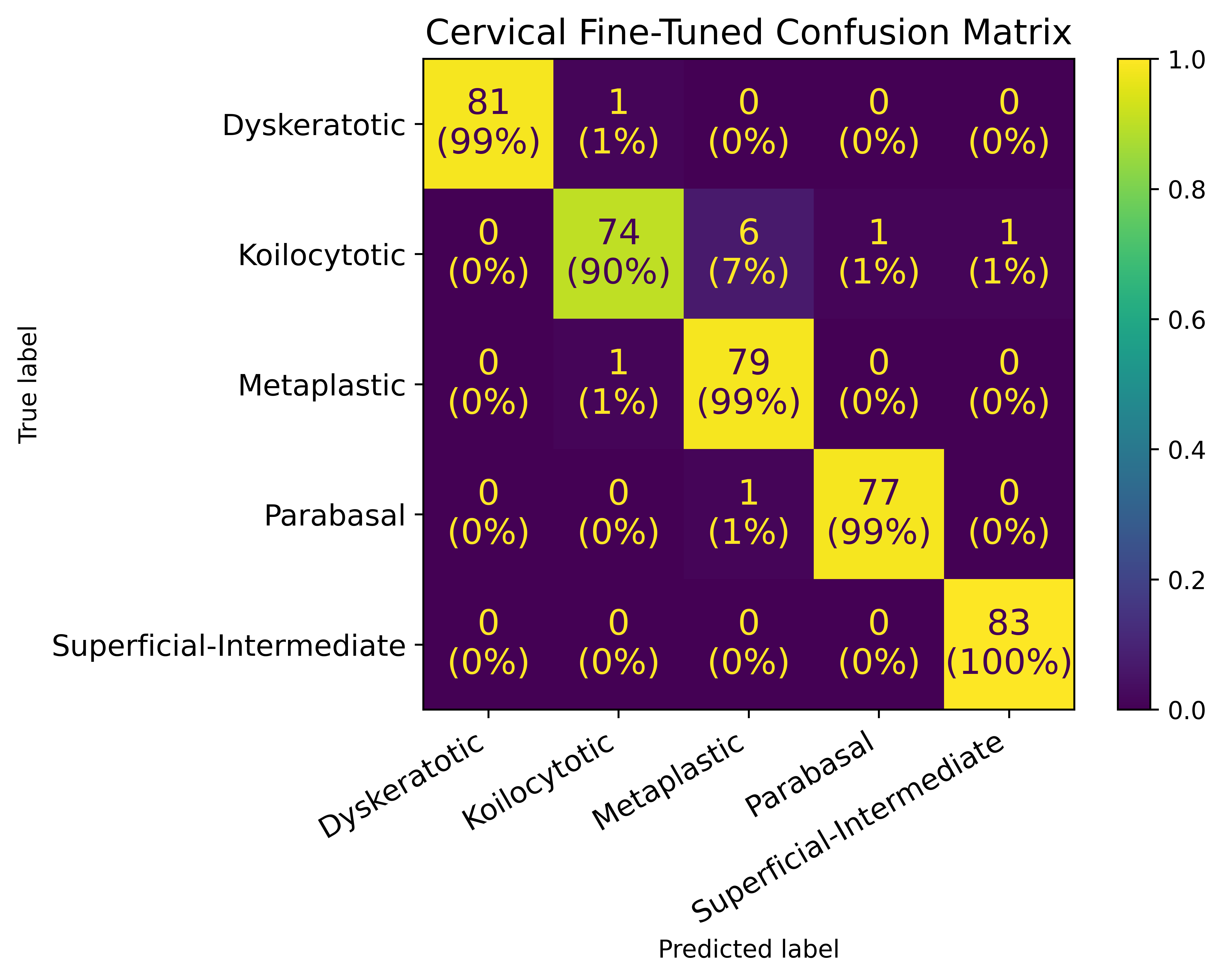}
    \caption{Confusion matrix for cervical cell classification using hybrid classical-quantum approach.}
    \label{fig:cm_cervical}
\end{figure}

Figure~\ref{fig:cm_cervical} shows the confusion matrix of the hybrid quantum-classical fine-tuned model for cervical cell classification. The quantum-based model correctly classifies 99\% of Dyskeratotic, 90\% of Koilocytotic, 99\% of Metaplastic, 99\% of Parabasal, and 100\% of Superficial-Intermediate cases, with errors of 1 Dyskeratotic $\rightarrow$ Koilocytotic, 6 Koilocytotic $\rightarrow$ Metaplastic, 1 Koilocytotic $\rightarrow$ Parabasal, 1 Koilocytotic $\rightarrow$ Superficial-Intermediate, 1 Metaplastic $\rightarrow$ Koilocytotic, and 1 Parabasal $\rightarrow$ Metaplastic.

\begin{figure}[tbh!]
        \centering
        \includegraphics[width=\linewidth]{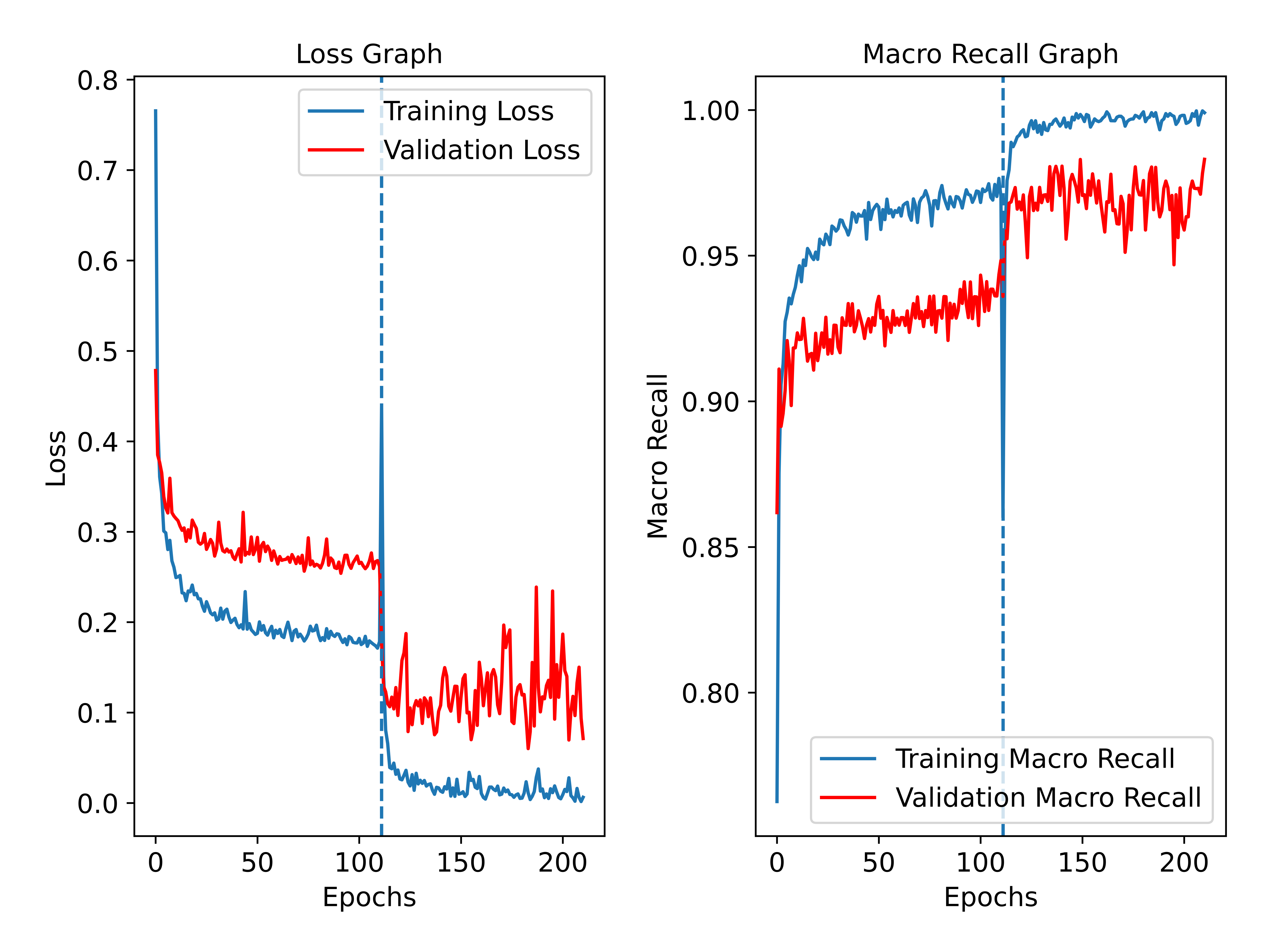}
    \caption{Training and validation loss and macro recall for cervical cell classification. The dashed vertical line marks the transition point as a result of fine-tuning.}
    \label{fig:curves_cervical}
\end{figure}

Figure~\ref{fig:curves_cervical} shows that cervical cell classification using the quantum-based method reaches high macro recall. The hybrid model follows a smooth trajectory. Before the dashed line, both training and validation loss decrease gradually while macro recall increases steadily; after the dashed line (fine-tuning), the training loss drops further and remains very low, while the validation loss also decreases overall with some fluctuations. At the same time, both training and validation macro recall improve after fine-tuning and then stabilize at a high level.

Table~\ref{tab:cervical_metrics} compares the class-wise and aggregate metrics for quantum and classical models. The hybrid quantum model has higher Macro Precision, Macro Recall, Macro F1, Micro Precision, Micro Recall, Micro F1, and MCC. The hybrid quantum model has higher Dyskeratotic Precision, Dyskeratotic Recall, Dyskeratotic F1 score, Koilocytotic Precision, Koilocytotic Recall, Koilocytotic F1 score, Metaplastic Recall, and Metaplastic F1 score. The classical CNN has slightly higher Parabasal Recall (1.0000 vs. 0.9872), Parabasal F1 score (0.9936 vs. 0.9872). The quantum model is better overall.
\begin{table}[htb!]
\caption{Performance comparison between the hybrid quantum-classical and classical CNN fine-tuned models for cervical cell classification.}
\label{tab:cervical_metrics}
\centering
\hspace*{-0.4\columnwidth}
\setlength{\tabcolsep}{2.5pt}
\renewcommand{\arraystretch}{0.95}
\begin{minipage}[t]{0.5\columnwidth}\centering
\begin{tabular}{lcc}
\hline
\textbf{Metric} & \textbf{Hybrid QC} & \textbf{Classical CNN} \\
\hline
Dyskeratotic Precision            & 1.0000 & 0.9756 \\
Dyskeratotic Recall               & 0.9878 & 0.9756 \\
Dyskeratotic F1 Score             & 0.9939 & 0.9756 \\
\hline
Koilocytotic Precision            & 0.9737 & 0.9600 \\
Koilocytotic Recall               & 0.9024 & 0.8780 \\
Koilocytotic F1 Score             & 0.9367 & 0.9172 \\
\hline
Metaplastic Precision             & 0.9186 & 0.9176 \\
Metaplastic Recall                & 0.9875 & 0.9750 \\
Metaplastic F1 Score              & 0.9518 & 0.9455 \\
\hline
Parabasal Precision               & 0.9872 & 0.9873 \\
Parabasal Recall                  & 0.9872 & 1.0000 \\
Parabasal F1 Score                & 0.9872 & 0.9936 \\
\hline
Superficial-Intermediate Precision & 0.9881 & 0.9881 \\
Superficial-Intermediate Recall    & 1.0000 & 1.0000 \\
Superficial-Intermediate F1 Score  & 0.9940 & 0.9940 \\
\hline
Macro Precision                   & 0.9735 & 0.9657 \\
Micro Precision                   & 0.9728 & 0.9654 \\
Macro Recall                      & 0.9730 & 0.9657 \\
Micro Recall                      & 0.9728 & 0.9654 \\
Macro F1                          & 0.9727 & 0.9652 \\
Micro F1                          & 0.9728 & 0.9654 \\
MCC                               & 0.9663 & 0.9571 \\
Test Accuracy                     & 0.97   & 0.97   \\
\hline
\end{tabular}
\end{minipage}
\vspace{-0.4em}
\end{table}

Next, we explain our dataset-specific customizations to diagnose brain tumor leveraging quantum power. 

\section{Brain Tumor Image Classification}
\label{sec:brain}

\begin{figure}[tbh!]
    \centering
    \includegraphics[width=.5\textwidth]{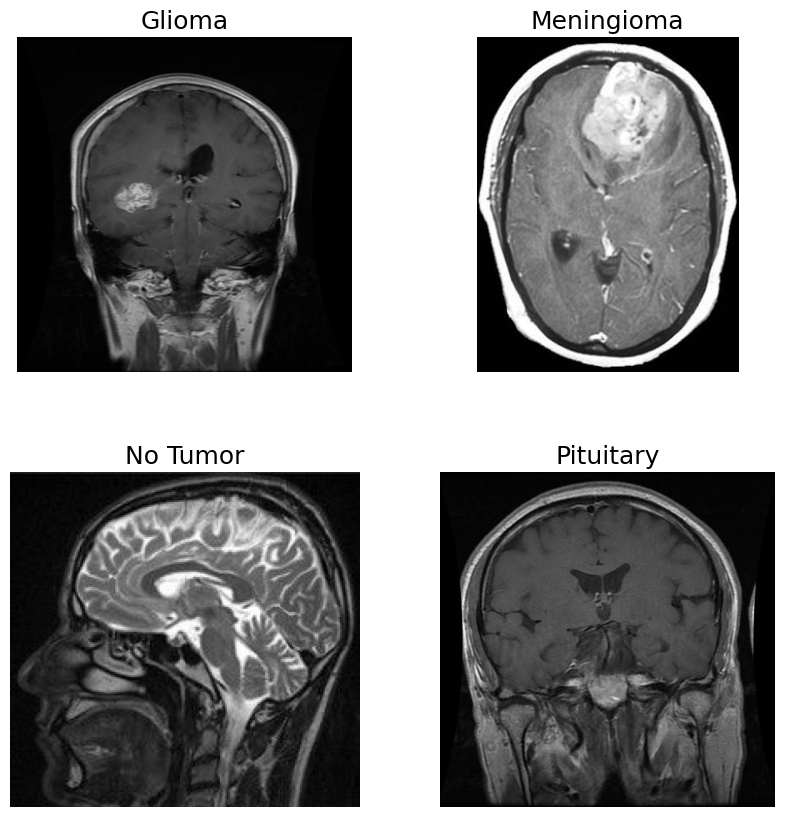}
    \caption{Sample images from brain tumor dataset~\cite{ghaffar2024} displaying each of the four classes: Glioma, Meningioma, No Tumor, and Pituitary.}
    \label{fig:brain_class_imgs}
\end{figure}

We apply QCNNs to brain MRI images \cite{ghaffar2024} to classify different types of tumors such as glioma, meningioma, pituitary, or no tumors. The dataset consists of $7023$ total images; glioma with $1621$ images, meningioma with $1645$ images, no tumor with $2000$ images, and pituitary with $1757$ images. An image of each class is shown in Figure~\ref{fig:brain_class_imgs}.

We split the dataset into three parts: $80\%$ for training, $10\%$ for validation, and $10\%$ for testing. When loading each image through the PyTorch Dataset subclass, we apply a preprocessing function to each image similarly done in the kidney dataset with CLAHE and denoising. Due to the brain MRI images containing more complex features, we reduce the filter strength during denoising to preserve more image details. Data augmentation techniques such as random cropping, rotations, and vertical and horizontal flips are also utilized to introduce more variety along with a preprocessing function for the EfficientNetB$4$ model.  

For the feature extractor, we use the pretrained EfficientNetB$4$ model. The weights of EfficientNetB$4$ were frozen before training and the classification head was changed to be a fully connected \texttt{Linear} layer that reduced from $1792$ features to $256$ features.  

\begin{figure}[htbp!]
    \centering
\includegraphics[width=\linewidth,height=0.42\textheight,keepaspectratio]{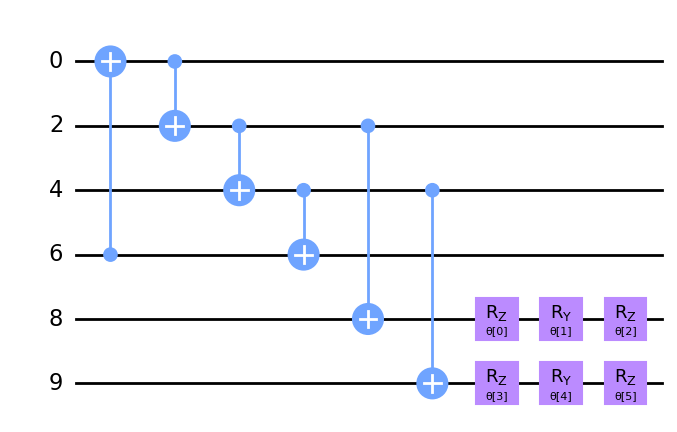}
    \caption{The third interaction layer for brain diagnosis. Unlike the third interaction layer in~\cite{mahmud2024quantum}, which assigns one ancilla qubit per class, the proposed layer uses the joint superposed state of the ancilla register to encode multiple class states, allowing a maximum of $4$ output states with $2$ ancilla qubits.}
    \label{fig:brain_third_interact}
\end{figure}

Amplitude encoding was used as the quantum encoding method. We used a total of $8$ qubits as input from the $256$ features. The QCNN model contains a total of $5$ quantum layers, similar to Section~\ref{sec:cervic}. However, the $3$rd interaction layer is adjusted to have $2$ ancilla qubits. A diagram of this change is shown in Figure~\ref{fig:brain_third_interact}.

We measure the probability values for the ancilla register to get a total of $4$ possible quantum states. The quantum states were then converted to logits to match the output of the cross entropy loss function. 
To improve training stability, the QCNN parameters were initialized from a uniform distribution over $[-1,1]$. We additionally applied gradient clipping with a maximum norm of $1.0$ during optimization. These choices were used as practical training heuristics.

Algorithm~\ref{alg:hybrid_qcnn_brainmri} summarizes the steps. 
\begin{algorithm*}[tbh!]
\setcounter{AlgoLine}{0}
\caption{Brain Tumor Classification (10 qubits: 8 data $q_0\!:\!q_7$ + 2 ancilla $a_0,a_1$)}
\label{alg:hybrid_qcnn_brainmri}
\DontPrintSemicolon
\footnotesize

\KwIn{Image $x$; class set $\{0,1,2,3\}$}
\KwOut{$\hat{\mathbf{y}}\in\mathbb{R}^4$}
\BlankLine

\textbf{Enhance:}
grayscale $\rightarrow$ fast NLM $(7,7,21)$ $\rightarrow$ CLAHE (clip$=5$) $+30$ shift $\rightarrow$ clip to $[0,255]$.\;

\textbf{Encode:}
$\mathbf{z}\leftarrow$$\mathrm{Enc}(x)\in\mathbb{R}^{256}$ (EfficientNetB4).\;

 \textbf{Amplitude Embedding:}
embed $\mathbf{z}$ into $8$ qubits via amplitude encoding.\;

 \textbf{Convolution Layer 1 ($15$ params):}
apply the convolutional circuit~9 of~\cite{Hur2022} on
$(q_0,q_7)$, then $(q_0,q_1),(q_2,q_3),(q_4,q_5),(q_6,q_7)$,
then $(q_1,q_2),(q_3,q_4),(q_5,q_6)$.\;

\textbf{Pooling ($2$ params; \cite{Hur2022}):}
for $i\in\{0,2,4,6\}$ apply 
$\mathrm{CRZ}(\phi_0)$–$X$–$\mathrm{CRX}(\phi_1)$ 
with control $q_{i+1}$ and target $q_i$;
retain $q_0,q_2,q_4,q_6$.\;

\textbf{Interaction Layer 1 (paramless; \cite{mahmud2024quantum}):}
apply Toffoli cascade on $\{q_0,q_2,q_4,q_6\}$.\;

\textbf{Convolution Layer 2 ($15$ params):}
on active wires $\{q_0,q_2,q_4,q_6\}$ apply the same convolutional circuit~9 of~\cite{Hur2022} on
$(q_0,q_6)$, $(q_0,q_2)$, $(q_4,q_6)$, $(q_2,q_4)$.\;

\textbf{Interaction Layer 2 ($12$ params; \cite{mahmud2024quantum}):}
apply Toffoli–Toffoli–$R_X$–Toffoli–$R_Y$–Toffoli–$R_Z$
on $\{q_0,q_2,q_4,q_6\}$.\;

\textbf{Classifier Interaction ($6$ params):}
CNOT ring 
$q_6{\rightarrow}q_0{\rightarrow}q_2{\rightarrow}q_4{\rightarrow}q_6$;
then
$\mathrm{CNOT}(q_2{\rightarrow}a_0)$,
$\mathrm{CNOT}(q_4{\rightarrow}a_1)$;
apply $R_ZR_YR_Z$ rotations on each 2 ancilla for 4 classes.\;

\textbf{Measurement:}
$\mathbf{p}=$$\mathrm{Prob}(a_0,a_1)\in\mathbb{R}^4$.\;

\textbf{Logits:}
$\hat{\mathbf{y}}=\log(\mathbf{p}+\epsilon)$, $\epsilon=10^{-10}$.\;

\BlankLine
\textbf{Initialization:} initialize pretrained EfficientNetB4 and freeze encoder parameters;\;
\textbf{Pre-fine-tuning:} train the hybrid model with Adam and cross-entropy loss for $200$ epochs to obtain a pre-fine-tuned checkpoint;\;
\textbf{Fine-tuning setup:} load pre-fine-tuned checkpoint weights;\;
unfreeze last $N\in\{1,\dots,4\}$ encoder blocks;\;
\textbf{Optimize:} Adam (encoder lr $=\eta_e$, QCNN+classifier lr $=\eta$) 
minimize cross-entropy loss for $100$ epochs;\;
\textbf{Train \& select model:} apply gradient clipping (max-norm $1.0$) and keep frozen BatchNorm layers in eval mode;\;
track validation macro-recall and save checkpoint.\;
\end{algorithm*}

The hybrid QCNN model was trained for $200$ epochs. We prevent the mean and variance of the batch normalization layers from updating during training by keeping them in evaluation mode. This helped improve performance metrics as it did not update its weights to adjust to the dataset and only outputted general features that the QCNN model was able to learn deeper. We used the Adam optimizer and cross entropy loss. Using Optuna \cite{akiba2019}, the search space for learning rate was between $0.1$ and $0.00001$ while batch size was between $8$ and $64$. We trained the hybrid QCNN model over $10$ different trials and focused on maximizing macro recall on the validation set.

\subsection{Results and Analysis on Brain Tumor Images}

\begin{figure}[tbh!]
        \centering
        \includegraphics[width=\linewidth]{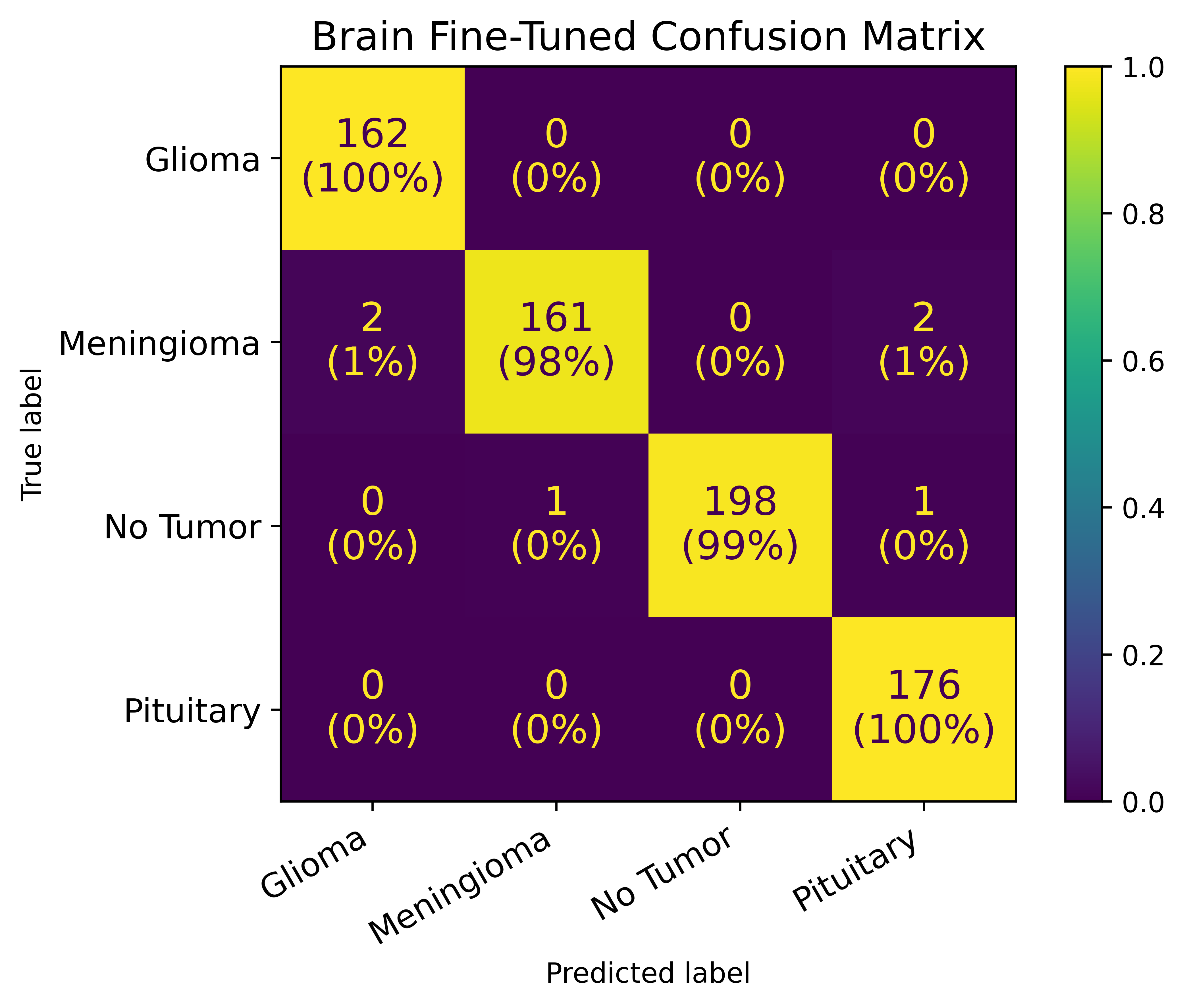}
    \caption{Confusion matrix for brain tumor classification using hybrid classical-quantum approach.}
    \label{fig:cm_brain}
\end{figure}

Figure~\ref{fig:cm_brain} shows the confusion matrix of the hybrid quantum-classical fine-tuned model for brain tumor classification. The quantum-based model correctly classifies 100\% of Glioma, 98\% of Meningioma, 99\% of No Tumor, and 100\% of Pituitary cases, with errors of 2 Meningioma $\rightarrow$ Glioma, 2 Meningioma $\rightarrow$ Pituitary, 1 No Tumor $\rightarrow$ Meningioma, and 1 No Tumor $\rightarrow$ Pituitary. 

\begin{figure}[tbh!]
        \centering
        \includegraphics[width=\linewidth]{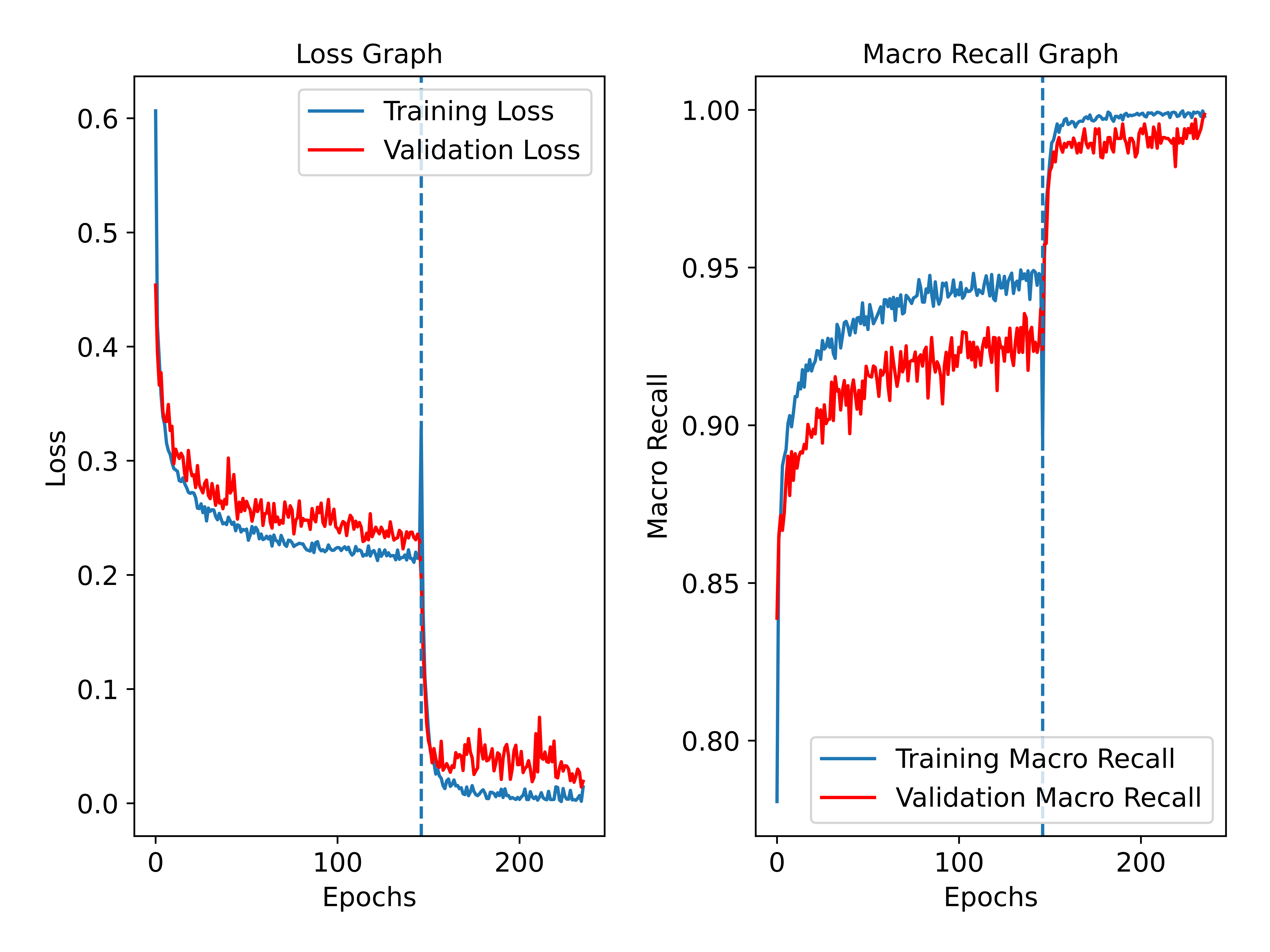}
    \caption{Training and validation loss and macro recall for brain tumor classification. The dashed vertical line marks the transition point as a result of fine-tuning.}
    \label{fig:curves_brain}
\end{figure}

Figure~\ref{fig:curves_brain} shows that the brain disease classification using the quantum-based method reaches high macro recall. The hybrid model follows a smooth trajectory. Before the dashed line, both training and validation loss decrease gradually while macro recall increases steadily; after the dashed line (fine-tuning), both losses drop sharply to very small values and both macro recall curves stabilize close to 0.99. 

Table~\ref{tab:brain_metrics} compares the class-wise and aggregate metrics for quantum and classical models. The hybrid quantum model has higher Macro Precision, Macro Recall, Macro F1, Micro Precision, Micro Recall, Micro F1, and MCC. The hybrid quantum model has also higher Glioma Recall, Glioma F1 score, Meningioma Precision, No Tumor Recall, No Tumor F1 score, Pituitary Precision, and Pituitary F1 score. No Tumor Precision is 1.0000 for both models, and Pituitary Recall is 1.0000 for both models. The quantum model is better overall.
\begin{table}[htb!]
\caption{Performance comparison between the hybrid quantum-classical and classical CNN fine-tuned models for brain tumor classification.}
\label{tab:brain_metrics}
\centering
\hspace*{-0.4\columnwidth}
\setlength{\tabcolsep}{2.5pt}
\renewcommand{\arraystretch}{0.95}
\begin{minipage}[t]{0.5\columnwidth}\centering
\begin{tabular}{lcc}
\hline
\textbf{Metric} & \textbf{Hybrid QC} & \textbf{Classical CNN} \\
\hline
Glioma Precision      & 0.9878 & 0.9938 \\
Glioma Recall         & 1.0000 & 0.9877 \\
Glioma F1 Score       & 0.9939 & 0.9907 \\
\hline
Meningioma Precision  & 0.9938 & 0.9879 \\
Meningioma Recall     & 0.9758 & 0.9879 \\
Meningioma F1 Score   & 0.9847 & 0.9879 \\
\hline
No Tumor Precision    & 1.0000 & 1.0000 \\
No Tumor Recall       & 0.9900 & 0.9800 \\
No Tumor F1 Score     & 0.9950 & 0.9899 \\
\hline
Pituitary Precision   & 0.9832 & 0.9724 \\
Pituitary Recall      & 1.0000 & 1.0000 \\
Pituitary F1 Score    & 0.9915 & 0.9860 \\
\hline
Macro Precision       & 0.9912 & 0.9885 \\
Micro Precision       & 0.9915 & 0.9886 \\
Macro Recall          & 0.9914 & 0.9889 \\
Micro Recall          & 0.9915 & 0.9886 \\
Macro F1              & 0.9913 & 0.9886 \\
Micro F1              & 0.9915 & 0.9886 \\
MCC                   & 0.9886 & 0.9848 \\
Test Accuracy         & 0.99   & 0.99   \\
\hline
\end{tabular}
\end{minipage}
\vspace{-0.4em}
\end{table}

\section{Conclusion}
\label{sec:conclus}

In this work, we leveraged hybrid classical-quantum models for robust multi-class medical image diagnosis across three distinct imaging tasks: kidney disease classification from CT images, cervical cell classification from pap smear images, and brain tumor classification from MRI scans. By combining dataset-specific preprocessing, transfer learning with pretrained encoders, and quantum convolutional neural networks, the proposed framework was designed to improve feature representation while maintaining a compact set of trainable parameters.

The experimental results demonstrate that the proposed hybrid models achieve strong and stable performance across all datasets. For kidney CT classification, both the $8$-qubit and $12$-qubit models achieved high accuracy, with the $8$-qubit configuration showing slightly better robustness in clinically important classes such as Stone and Tumor. For cervical cell classification and brain tumor classification, the modified hybrid QCNN architectures incorporating probabilistic ancilla-state measurement enabled effective multi-class prediction and consistently outperformed corresponding classical CNN baselines under comparable encoder and training settings. Fine-tuning the pretrained encoders further improved convergence behavior and classification performance, highlighting the benefit of adapting classical feature extractors to domain-specific medical data before quantum processing.

The use of probabilistic states for the cervical and brain classification tasks helped represent a greater number of classes, allowing the QCNN to more effectively leverage quantum superposition for multi-class classification tasks. By reducing the learning rate and unfreezing parts of the encoder, the $11$-qubit cervical and $10$-qubit brain hybrid models were able to adapt their trainable parameters to their own respective datasets, allowing the QCNN to extract meaningful features. 

These findings suggest that coupling robust transfer learning classical feature extraction with quantum neural architectures provides diagnostic capability in complex medical datasets. The results suggest that quantum-inspired diagnostic pipelines may offer meaningful advantages in accuracy, parameter efficiency, and scalability for complex medical imaging tasks. Future work will investigate validation on additional clinical datasets and robustness under domain shift to further assess practical diagnostic utility.

\bibliographystyle{IEEEtran}
\bibliography{refs}

@article{mahmud2024quantum,
  title={Quantum convolutional neural networks with interaction layers for classification of classical data},
  author={Mahmud, Jishnu and Mashtura, Raisa and Fattah, Shaikh Anowarul and Saquib, Mohammad},
  journal={Quantum Machine Intelligence},
  volume={6},
  number={1},
  pages={11},
  year={2024},
  publisher={Springer},
  doi={10.1007/s42484-024-00145-4},
  url={https://doi.org/10.1007/s42484-024-00145-4}
}

@article{cong,
  title={Quantum convolutional neural networks},
  author={Cong, Iris and Choi, Soonwon and Lukin, Mikhail D},
  journal={Nature Physics},
  volume={15},
  number={12},
  pages={1273--1278},
  year={2019},
  publisher={Nature Publishing Group UK London},
  doi = {10.1038/s41567-019-0648-8},
  url = {https://doi.org/10.1038/s41567-019-0648-8}
}

@article{kaggle_kidney,
    title={{CT} KIDNEY DATASET: Normal-Cyst-Tumor and Stone},
    author={Md Nazmul Islam},
    howpublished={Kaggle},
    year={2021},
    note={[Online]. Available: \url{https://www.kaggle.com/datasets/nazmul0087/ct-kidney-dataset-normal-cyst-tumor-and-stone}}
    }

@incollection{clahe,
  title={Contrast limited adaptive histogram equalization},
  author={Zuiderveld, Karel},
  booktitle={Graphics gems IV},
  pages={474--485},
  year={1994}
}

@INPROCEEDINGS{tantawi2023brainmri,
  author={Tantawi, Baraa and Ahmed, Hamza Kamel and Magdy, Malak and Adel, Mohamed and Sayed, Gehad Ismail},
  booktitle={2023 Intelligent Methods, Systems, and Applications (IMSA)}, 
  title={Exploring the Power of Quantum Convolutional Neural Networks for Brain {MRI} Image Classification}, 
  year={2023},
  volume={},
  number={},
  pages={549-584},
  doi={10.1109/IMSA58542.2023.10217624},
  url={https://doi.org/10.1109/IMSA58542.2023.10217624}
}

@inproceedings{stalin2025dr,
  author={Babu, G Stalin and Lakshmi Sathwika, M. and Yamuna, V and Sahithi, K and Rohith, N and Abhishek, P},
  booktitle={4th OPJU International Technology Conference (OTCON) on Smart Computing for Innovation and Advancement in Industry 5.0}, 
  title={Hybrid Quantum Convolutional Neural Networks for Enhanced Diabetic Retinopathy Detection}, 
  year={2025},
  volume={},
  number={},
  pages={1-6},
  doi= {10.1109/OTCON65728.2025.11070674},
  url = {https://doi.org/10.1109/OTCON65728.2025.11070674},
  publisher = {IEEE}
}

@article{xiang2024qccnn,
  title={Quantum-classical hybrid convolutional neural networks for breast cancer diagnosis},
  author={Xiang, Qiuyu and Li, Dongfen and Hu, Zhikang and Yuan, Yuhang and Sun, Yuchen and Zhu, Yonghao and Fu, You and Jiang, Yangyang and Hua, Xiaoyu},
  journal={Scientific Reports},
  volume={14},
  pages={24699},
  year={2024},
  publisher={Nature Publishing Group},
  doi = {10.1038/s41598-024-74778-7},
  url = {https://doi.org/10.1038/s41598-024-74778-7}
}

@inproceedings{khatoniar2024hybrid,
  author={Khatoniar, Ria and Vyas, Vishu and Acharya, Videet and Saxena, Anshul and Saxena, Amit and Neiwal, Rahul and Korgaonkar, Kunal},
  booktitle={2024 International Conference on Trends in Quantum Computing and Emerging Business Technologies}, 
  title={Quantum Convolutional Neural Network for Medical Image Classification: A Hybrid Model}, 
  year={2024},
  volume={},
  number={},
  pages={1-5},
  doi={10.1109/TQCEBT59414.2024.10545121},
  url = {https://doi.org/10.1109/TQCEBT59414.2024.10545121},
  publisher ={IEEE}
}

@inproceedings{ovi2025multiqubit,
  author={Ovi, Tareque Bashar and Bashree, Nomaiya and Alam, Ayat Subah and Tanzim, Rawnak and Wahed, Md Abdul and Nyeem, Hussain},
  booktitle={2025 International Conference on Quantum Photonics, Artificial Intelligence, and Networking (QPAIN)}, 
  title={Multiqubit Quantum Convolutional Neural Networks for Efficient {AI}-Driven Healthcare Analytics}, 
  year={2025},
  volume={},
  number={},
  pages={1-6},
  doi={10.1109/QPAIN66474.2025.11172247},
  url = {https://doi.org/10.1109/QPAIN66474.2025.11172247},
  publisher = {IEEE}
}

@INPROCEEDINGS{cruz2025quanvolution,
  author={Cruz, Meenaloshini Vimal and Dhar, Sourabh and Hajra, Samriddha and Usha, G. and Gautam, Kumar},
  booktitle={2025 6th International Conference on Inventive Research in Computing Applications (ICIRCA)}, 
  title={Quantum Image Classification for Sustainable Medical Image Diagnosis}, 
  year={2025},
  volume={},
  number={},
  pages={1789-1792},
  doi={10.1109/ICIRCA65293.2025.11089792}}

@inproceedings{li2025medicalmnist,
title = {A quantum convolutional network and {ResNet} (50)-based classification architecture for the {MNIST} medical dataset},
journal = {Biomedical Signal Processing and Control},
volume = {87},
pages = {105560},
year = {2024},
issn = {1746-8094},
doi = {https://doi.org/10.1016/j.bspc.2023.105560},
url = {https://www.sciencedirect.com/science/article/pii/S174680942300993X},
author = {Esraa Hassan and M. Shamim Hossain and Abeer Saber and Samir Elmougy and Ahmed Ghoneim and Ghulam Muhammad},
}

@INPROCEEDINGS{resnet,
  author={He, Kaiming and Zhang, Xiangyu and Ren, Shaoqing and Sun, Jian},
  booktitle={2016 IEEE Conference on Computer Vision and Pattern Recognition (CVPR)}, 
  title={Deep Residual Learning for Image Recognition}, 
  year={2016},
  volume={},
  number={},
  pages={770-778},
  doi={10.1109/CVPR.2016.90}}

@article{Hur2022,
  author    = {Tak Hur and Leeseok Kim and Daniel K. Park},
  title     = {Quantum convolutional neural network for classical data classification},
  journal   = {Quantum Machine Intelligence},
  year      = {2022},
  volume    = {4},
  number    = {1},
  pages     = {3},
  doi       = {10.1007/s42484-021-00061-x},
  url       = {https://doi.org/10.1007/s42484-021-00061-x},
  issn      = {2524-4914}
}

@INPROCEEDINGS{Baral2024,
  author={Baral, Biswaraj and Roy, Taposh Dutta},
  booktitle={2024 4th International Conference on Sustainable Expert Systems (ICSES)}, 
  title={Integrating Quantum Principles with Classical Transfer Learning Approaches for Pneumonia Detection from Chest Radiographs}, 
  year={2024},
  volume={},
  number={},
  pages={1042-1049},
  doi={10.1109/ICSES63445.2024.10763087},
  url={https://doi.org/10.1109/ICSES63445.2024.10763087}
}

@INPROCEEDINGS{Kulkarni2025,
  author={Kulkarni, Savitri and V, Shashidhar and Shenoy, P. Deepa and R, Venugopal K.},
  booktitle={2025 IEEE International Conference for Women in Innovation, Technology \& Entrepreneurship (ICWITE)}, 
  title={Quantum Convolutional Neural Network for Coffee Plant Disease Diagnosis}, 
  year={2025},
  volume={},
  number={},
  pages={1-6},
  doi={10.1109/ICWITE64848.2025.11307161},
  url={https://doi.org/10.1109/ICWITE64848.2025.11307161}
}

@INPROCEEDINGS{Chandra2022,
  author={Chandra, Sanskriti and Saxena, Shivam and Kumar, Santosh and Chaube, Mithilesh Kumar and Bodhey, Narendra Kuber},
  booktitle={2022 IEEE International Women in Engineering (WIE) Conference on Electrical and Computer Engineering (WIECON-ECE)}, 
  title={A Novel Framework For Brain Disease Classification Using Quantum Convolutional Neural Network}, 
  year={2022},
  volume={},
  number={},
  pages={346-351},
  doi={10.1109/WIECON-ECE57977.2022.10150851},
  url={https://doi.org/10.1109/WIECON-ECE57977.2022.10150851}
}

@ARTICLE{pravin2023,
  author={Pravin, Sheena Christabel and Rohith, G. and Kiruthika, V. and Manikandan, E. and Methelesh, S. and Manoj, A.},
  journal={IEEE Access}, 
  title={Underwater Animal Identification and Classification Using a Hybrid Classical-Quantum Algorithm}, 
  year={2023},
  volume={11},
  number={},
  pages={141902-141914},
  doi={10.1109/ACCESS.2023.3343120},
  url={https://doi.org/10.1109/ACCESS.2023.3343120}
}

@article{ajlouni2023,
    author = {Ajlouni, Naim and Ozyavas, Adem and Takaoglu, Mustafa and Takaoglu, Faruk and Ajlouni, Firas},
    title = {Medical image diagnosis based on adaptive Hybrid Quantum {CNN}},
    journal = {BMC Medical Imaging},
    year = {2023},
    volume={23},
    doi = {10.1186/s12880-023-01084-5},
    url = {https://doi.org/10.1186/s12880-023-01084-5}
}

@article{rao2024,
    title = {Hybrid framework for respiratory lung diseases detection based on classical {CNN} and quantum classifiers from chest {X-rays}},
    journal = {Biomedical Signal Processing and Control},
    volume = {88},
    pages = {105567},
    year = {2024},
    issn = {1746-8094},
    doi = {https://doi.org/10.1016/j.bspc.2023.105567},
    url = {https://www.sciencedirect.com/science/article/pii/S1746809423010005},
    author = {G.V. Eswara Rao and Rajitha B. and Parvathaneni Naga Srinivasu and Muhammad Fazal Ijaz and Marcin Woźniak}
}

@article{alfajri2026,
    title = {Hybrid quantum-classical convolutional neural networks for multiclass lung {X-ray} image classification},
    journal = {Journal of Radiation Research and Applied Sciences},
    volume = {19},
    number = {1},
    pages = {102165},
    year = {2026},
    issn = {1687-8507},
    doi = {https://doi.org/10.1016/j.jrras.2026.102165},
    url = {https://www.sciencedirect.com/science/article/pii/S1687850726000087},
    author = {Ariiq Islam Alfajri and Tony Sumaryada and Sitti Yani and Duong Thanh Tai and Nissren Tamam and Abdelmoneim Sulieman}
}

@article{li2025,
    title = {Hybrid classical–-quantum neural networks enhanced by quantum architecture search for coronary artery stenosis detection},
    journal = {Neurocomputing},
    volume = {618},
    pages = {129111},
    year = {2025},
    issn = {0925-2312},
    doi = {https://doi.org/10.1016/j.neucom.2024.129111},
    url = {https://www.sciencedirect.com/science/article/pii/S0925231224018824},
    author = {Shaochun Li and Junzhi Cui and Jingli Ren}
}

@article{frasca2025,
    title = {Optimizing melanoma diagnosis: A hybrid deep learning and quantum computing approach for enhanced lesion classification},
    journal = {Intelligence-Based Medicine},
    volume = {12},
    pages = {100264},
    year = {2025},
    issn = {2666-5212},
    doi = {https://doi.org/10.1016/j.ibmed.2025.100264},
    url = {https://www.sciencedirect.com/science/article/pii/S2666521225000687},
    author = {Maria Frasca and Ilaria Cutica and Gabriella Pravettoni and Davide {La Torre}}
}

@article{xiao2025,
  title={Complex-encoded quantum convolutional neural networks},
  author={Xiao, Zhipeng and Tan, Xiaoqing and Bao, Daipengwei and Huang, Rui},
  journal={Quantum Machine Intelligence},
  volume={7},
  number={84},
  year={2025},
  doi={10.1007/s42484-025-00310-3},
  url={https://doi.org/10.1007/s42484-025-00310-3}
}

@INPROCEEDINGS{plissiti2018,
  author={Plissiti, Marina E. and Dimitrakopoulos, P. and Sfikas, G. and Nikou, Christophoros and Krikoni, O. and Charchanti, A.},
  booktitle={2018 25th IEEE International Conference on Image Processing (ICIP)}, 
  title={Sipakmed: A New Dataset for Feature and Image Based Classification of Normal and Pathological Cervical Cells in Pap Smear Images}, 
  year={2018},
  volume={},
  number={},
  pages={3144-3148},
  doi={10.1109/ICIP.2018.8451588}}

@ARTICLE{yousif2024,
  author={Yousif, Mohammed and Al-Khateeb, Belal and Garcia-Zapirain, Begonya},
  journal={IEEE Access}, 
  title={A New Quantum Circuits of Quantum Convolutional Neural Network for {X-ray} Images Classification}, 
  year={2024},
  volume={12},
  number={},
  pages={65660-65671},
  doi={10.1109/ACCESS.2024.3396411},
  url={https://doi.org/10.1109/ACCESS.2024.3396411}
}

@INPROCEEDINGS{arepalli2025,
  author={Arepalli, Peda Gopi and Chennamsetti, Sravya and Poluri, Mary Midhila and Kumpati, Sukanya and Pulugujju, Rathna Kumari and Karanki, Hemasri},
  booktitle={2025 7th International Conference on Intelligent Sustainable Systems (ICISS)}, 
  title={Corn Leaf Disease Detection using Quantum {CNN}}, 
  year={2025},
  volume={},
  number={},
  pages={1235-1239},
  doi={10.1109/ICISS63372.2025.11076546},
  url={https://doi.org/10.1109/ICISS63372.2025.11076546}
}

@INPROCEEDINGS{kim2023,
  author={Kim, Ryan},
  booktitle={2023 IEEE International Conference on Quantum Computing and Engineering (QCE)}, 
  title={Implementing a Hybrid Quantum-Classical Neural Network by Utilizing a Variational Quantum Circuit for Detection of Dementia}, 
  year={2023},
  volume={02},
  number={},
  pages={256-257},
  doi={10.1109/QCE57702.2023.10231},
  url={https://doi.org/10.1109/QCE57702.2023.10231}
}

@ARTICLE{zhu2024,
  author={Zhu, Yijie and Bouridane, Ahmed and Celebi, M Emre and Konar, Debanjan and Angelov, Plamen and Ni, Qiang and Jiang, Richard},
  journal={IEEE Transactions on Artificial Intelligence}, 
  title={Quantum Face Recognition With Multigate Quantum Convolutional Neural Network}, 
  year={2024},
  volume={5},
  number={12},
  pages={6330-6341},
  doi={10.1109/TAI.2024.3419077},
  url={https://doi.org/10.1109/TAI.2024.3419077}
}

@article{kotwal2025,
  title={Sensor Infused Quantum {CNN} for Diabetes Disease Prediction and Diet Recommendation},
  author={Kotwal, Jameer and Futane, Pravin and Chavan, Gurunath and Chaudhari, Archana and Jose, Jithina and Khan Vajid
  },
  journal={International Journal of Computational Intelligence Systems},
  volume={18},
  number={113},
  year={2025},
  publisher={Springer},
  doi={10.1007/s44196-025-00833-4},
  url={https://doi.org/10.1007/s44196-025-00833-4}
}

@article{pandey2025,
  title={A hybrid quantum–classical convolutional neural network with a quantum attention mechanism for skin cancer},
  author={Pandey, Pradyumn and Mandal, Shrabanti
  },
  journal={Scientific Reports},
  volume={16},
  number={1639},
  year={2025},
  publisher={Nature Publishing Group},
  doi={10.1038/s41598-025-31122-x},
  issn = {2045-2322},
  url={https://doi.org/10.1038/s41598-025-31122-x}
}

@article{radhika2025,
  title={Design and implementation of quantum hippo inspired convolutional neural networks using parametric quantum circuits for an efficient lung cancer classification},
  author={Radhika, S and Sharada, S
  },
  journal={Discover Computing},
  volume={28},
  number={122},
  year={2025},
  publisher={Springer},
  doi={10.1007/s10791-025-09634-x},
  url={https://doi.org/10.1007/s10791-025-09634-x}
}

@INPROCEEDINGS{kanimozhi2022,
  author={T, Kanimozhi and S, Sridevi and Manikumar, Tirumalanadhuni Siva and Dheeraj, Tallam and Sumanth, Amaravathi},
  booktitle={2022 International Conference on Innovative Trends in Information Technology (ICITIIT)}, 
  title={Brain Tumor Recognition based on Classical to Quantum Transfer Learning}, 
  year={2022},
  volume={},
  number={},
  pages={1-5},
  doi={10.1109/ICITIIT54346.2022.9744220},
  url={https://doi.org/10.1109/ICITIIT54346.2022.9744220}
}

@INPROCEEDINGS{vijayanand2025,
  author={Vijayanand, Sellamuthu Palanisamy and Tarafdar, Rajarshi and Priya, R M Senthil and Karthikayan, P.N. and Ganesan, Sivasathiya and S, Srimathi},
  booktitle={2025 Global Conference on Information Technology and Communication Networks (GITCON)}, 
  title={Deep Learning-Enabled Brain Tumor Classification Using Hybrid Quantum Convolutional Networks}, 
  year={2025},
  volume={},
  number={},
  pages={1-8},
  doi={10.1109/GITCON65266.2025.11376968},
  url={https://doi.org/10.1109/GITCON65266.2025.11376968}}

@MISC{ghaffar2024,
  title     = "Brain Tumor Data",
  author    = "Ghaffar, Ayesha",
  publisher = "Mendeley Data",
  year      =  "2024",
  doi = "10.17632/w4sw3s9f59.1",
  url = "https://doi.org/10.17632/w4sw3s9f59.1"
}

@ARTICLE{ahmad2025,
  author={Ahmad, Ishtiaq and Narmeen, Ramsha and Mughal, Umair Ahmad and Yang, Liang and Almadhor, Ahmad and Dhahbi, Sami and Wen, Miaowen and Ho, Pin-Han},
  journal={IEEE Wireless Communications}, 
  title={Quantum {CNN} for Detection and Identification of {UAV}-Enabled Non-Terrestrial Networks}, 
  year={2025},
  volume={32},
  number={3},
  pages={28-36},
  doi={10.1109/MWC.001.2400419},
  url={https://doi.org/10.1109/MWC.001.2400419}
}

@article{preskill2018,
  doi = {10.22331/q-2018-08-06-79},
  url = {https://doi.org/10.22331/q-2018-08-06-79},
  title = {Quantum {C}omputing in the {NISQ} era and beyond},
  author = {Preskill, John},
  journal = {{Quantum}},
  issn = {2521-327X},
  publisher = {{Verein zur F{\"{o}}rderung des Open Access Publizierens in den Quantenwissenschaften}},
  volume = {2},
  pages = {79},
  month = aug,
  year = {2018}
}

@inproceedings{akiba2019,
author = {Akiba, Takuya and Sano, Shotaro and Yanase, Toshihiko and Ohta, Takeru and Koyama, Masanori},
title = {Optuna: A Next-generation Hyperparameter Optimization Framework},
year = {2019},
isbn = {9781450362016},
publisher = {Association for Computing Machinery},
address = {New York, NY, USA},
url = {https://doi.org/10.1145/3292500.3330701},
doi = {10.1145/3292500.3330701},
booktitle = {Proceedings of the 25th ACM SIGKDD International Conference on Knowledge Discovery \& Data Mining},
pages = {2623–-2631},
numpages = {9},
location = {Anchorage, AK, USA},
series = {KDD '19}
}

@inproceedings{mordacci2024,
author = {Mordacci, Marco and Ferrari, Davide and Amoretti, Michele},
title = {Multi-Class Quantum Convolutional Neural Networks},
year = {2024},
isbn = {9798400706462},
publisher = {Association for Computing Machinery},
address = {New York, NY, USA},
url = {https://doi.org/10.1145/3660318.3660326},
doi = {10.1145/3660318.3660326},
booktitle = {Proceedings of the 2024 Workshop on Quantum Search and Information Retrieval},
pages = {9–-16},
numpages = {8},
location = {Pisa, Italy},
series = {QUASAR '24}
}

\end{document}